%% file: main.tex
\documentclass[10pt]{article} 
\usepackage[preprint]{tmlr}

\input{math_commands.tex}

\usepackage{hyperref}
\usepackage{url}

\usepackage{makecell}
\usepackage{multirow}
\usepackage{amsthm}
    \theoremstyle{plain}

    \theoremstyle{definition}
    \newtheorem{definition}{Definition}
    
    \newtheorem{property}{Property}
    \newtheorem{postulate}{Postulate}
    \theoremstyle{remark}

\usepackage{graphicx}
\usepackage{subcaption}   
\usepackage{caption}
\usepackage{booktabs}
\usepackage{amssymb}
\usepackage{bm}
\usepackage{bbm}

\title{TaylorPODA: A Taylor Expansion-Based Method to Improve Post-Hoc Attributions for Opaque Models}

\author{\name Yuchi Tang \email ytang87@sheffield.ac.uk \\
      \addr School of Electrical and Electronic Engineering\\
      University of Sheffield
      \AND
      \name Iñaki Esnaola \email esnaola@sheffield.ac.uk \\
      \addr School of Electrical and Electronic Engineering\\
      University of Sheffield
      \AND
      \name George Panoutsos \email g.panoutsos@sheffield.ac.uk \\
      \addr School of Electrical and Electronic Engineering\\
      University of Sheffiel}

\def\month{MM}  
\def\year{YYYY} 
\def\openreview{\url{https://openreview.net/forum?id=XXXX}} 

\begin{document}

\maketitle

\begin{abstract}
Post-hoc model-agnostic local attribution (LA) methods have been widely adopted to explain opaque AI models by quantifying feature-wise contributions. However, many existing methods rely on heuristic or only partially justified attribution mechanisms, while the quality of attribution itself is often shaped by downstream objectives without universally accepted standards. In this work, we propose Taylor exPansion-Originated aDaptive Attribution (TaylorPODA), a new post-hoc model-agnostic LA method grounded in the Taylor expansion framework. We first introduce a set of postulates, namely \textit{precision}, \textit{federation}, and \textit{zero-discrepancy}, which formalize principled requirements for explicitly and exhaustively attributing Taylor terms to the corresponding features. Based on these postulates, we analyze existing post-hoc model-agnostic LA methods and identify a fundamental tension between principled attribution and adaptation toward user-defined utilities. To address this challenge, TaylorPODA introduces a controllable allocation mechanism for Taylor interaction effects, enabling attribution results to adapt to downstream objectives while preserving the proposed postulates. Furthermore, although developed from a Taylor-expansion perspective, TaylorPODA also admits a Harsanyi-dividend interpretation, allowing the attribution mechanism to extend beyond model differentiability. Theoretical analysis demonstrates that TaylorPODA satisfies all the proposed postulates together with an additional \textit{adaptation} property. Empirical results across multiple datasets and both differentiable and non-differentiable models further show that TaylorPODA achieves consistently improved alignment with user-defined utilities while maintaining the communicability of the resulting explanations. Overall, this work provides a starting point toward more trustworthy XAI systems for the deployment of increasingly powerful yet opaque task models. The code is available at \url{https://github.com/Yuchi-TANG-Research/TaylorPODA}.
\end{abstract}

\section{Introduction}
\subsection{Background}

AI models have been increasingly adopted across a wide range of tasks. However, concerns regarding the trustworthiness of deploying these models remain significant~\citep{von2021transparency}, particularly in high-stakes domains such as healthcare~\citep{jin2022explainable}, finance~\citep{vcernevivciene2024explainable}, and energy~\citep{machlev2022explainable}. In many practical scenarios, users can only interact with an AI model through its input-output interface without access to its internal structure. Even when internal details are available, the underlying decision-making process may remain difficult to interpret. This challenge is especially prominent for many modern machine learning (ML) models with highly complex architectures, whose prediction mechanisms are often inherently opaque to human users. Consequently, there has been a growing demand for designing eXplainable AI (XAI) systems that support more trustworthy deployment of AI models, particularly as high-performing but opaque models increasingly surpass directly interpretable alternatives.

To address this challenge, a broad range of post-hoc XAI methods have been proposed to explain the behavior of already deployed opaque models. Among them, post-hoc model-agnostic methods have received substantial attention due to their portability across different model families. In contrast, model-specific explainability methods are often restricted to particular architectures. For example, Integrated Gradients~\citep{pmlr_v70_sundararajan17a} is designed for differentiable gradient-based models, whereas TreeSHAP~\citep{lundberg2020local} is tailored to tree-based models. Since no single model architecture consistently outperforms others across all tasks, explainability methods that remain applicable across heterogeneous models are particularly desirable.

Within post-hoc model-agnostic XAI, local attribution (LA) has emerged as one of the predominant paradigms. LA methods aim to quantify the contribution of each input feature toward a specific model prediction. Unlike global explanation approaches that characterize model behavior at the dataset level~\citep{fisher2019all,gregorutti2017correlation}, LA focuses on explaining individual predictions by analyzing feature-wise effects under a given input instance. Most existing post-hoc model-agnostic LA methods rely on repeatedly querying the model under different perturbation or masking conditions, thereby estimating the contribution of individual features through changes in the resulting outputs.

Despite the diversity of existing LA methods, many of them implicitly seek to marginalize the independent contribution of each feature under limited knowledge of the underlying model. To provide a unified analytical perspective, \cite{deng2024unifying} proposed a Taylor expansion framework for analyzing post-hoc model-agnostic LA methods. This framework reveals that the attribution mechanisms of many existing methods can be interpreted through the allocation of Taylor terms corresponding to independent and interaction effects among features.



\subsection{Related work}
Significant efforts have been made to enhance the explainability of opaque models. Early methods, such as partial dependence plots~\citep{friedman2001greedy} and individual conditional expectation plots~\citep{goldstein2015peeking}, visualize changes in model outputs as feature values are varied. Subsequently, more systematic post-hoc methods were developed. \cite{lime} introduced Local Interpretable Model-agnostic Explanations (LIME), which builds an interpretable local surrogate for the original model. In parallel, LA methods were developed to marginalize the contribution of individual features. Grounded in the Prediction Difference (Prediction Diff) approach~\citep{robnik2008explaining}, the Occlusion-based approach~\citep{zeiler2014visualizing} computes LA by masking the target features. Furthermore, \cite{vstrumbelj2014explaining} extended the Shapley value~\citep{shapley1953value} from cooperative game theory to the explanation of opaque
AI models. Building further on the Shapley value, \cite{lundberg2017unified} introduced SHapley Additive exPlanations (SHAP) and its widely adopted implementation, which inspired many variants \citep{frye2020asymmetric,aas2021explaining,watson2022rational}. 

Given the absence of ground truth in typical post-hoc model-agnostic settings, WeightedSHAP \citep{kwonWeightedSHAPAnalyzingImproving2022} relaxes SHAP's notion of local accuracy by adopting a more flexible semi-value formulation~\citep{dubey1977probabilistic,hart1989potential}. This relaxation enables the incorporation of “user-defined utilities”~\citep{kwonWeightedSHAPAnalyzingImproving2022}, allowing attribution results to be optimized according to task-specific notions of explanation goodness, such as the area under the prediction recovery error curve (AUP). More broadly, the incorporation of user-defined utilities echoes emerging perspectives in XAI that explanations are not “one-size-fits-all”, but should instead be tailored to the contextual needs \citep{dhanorkar2021needs}. 

Building on feature coalition-level attribution, \cite{sundararajan2020shapley} proposed the Shapley-Taylor Interaction Index, which assigns contribution credits to feature subsets rather than directly producing feature-wise attributions. Although this provides a comprehensive and fully enumerative characterization of feature interactions, it also shifts the explanation from a compact feature-wise representation to a potentially large collection of subset-level attribution values. This exponentially growing subset space can make the resulting explanation difficult to interpret and communicate.

\subsection{Our contributions}
Despite these advances, existing post-hoc model-agnostic LA methods still face an important challenge: achieving principled attribution mechanisms while retaining the flexibility required for downstream adaptation. Motivated by this gap, we develop Taylor exPansion-Originated aDaptive Attribution (TaylorPODA) and summarize the main contributions of this work as follows: 
\begin{itemize}
    \item \textbf{A postulate system for principled post-hoc model-agnostic LA methods.} We introduce three postulates, namely \textit{precision}, \textit{federation}, and \textit{zero-discrepancy}, which are grounded in the Taylor expansion framework. These postulates formalize the design principles for explicitly and exhaustively attributing the related Taylor terms to the involved features.

    \item \textbf{A new post-hoc model-agnostic method, TaylorPODA.} We propose TaylorPODA, which satisfies all the proposed postulates within the Taylor expansion framework. Moreover, TaylorPODA introduces an additional \textit{adaptation} property by flexibly allocating Taylor interaction effects, thereby enabling task-specific optimization via user-defined utility functions.

    \item \textbf{Comprehensive theoretical and empirical evaluation.} We theoretically analyze whether existing post-hoc model-agnostic LA methods satisfy or violate the proposed postulates, and demonstrate that TaylorPODA satisfies all the proposed postulates together with the additional \textit{adaptation} property. Furthermore, through empirical evaluation under multiple user-defined utility objectives, we demonstrate the effectiveness of TaylorPODA in achieving task-specific optimization while preserving a principled attribution mechanism. 

    
\end{itemize} 
\begin{figure}[h]
\centering
    \centerline{\includegraphics[width=\linewidth]{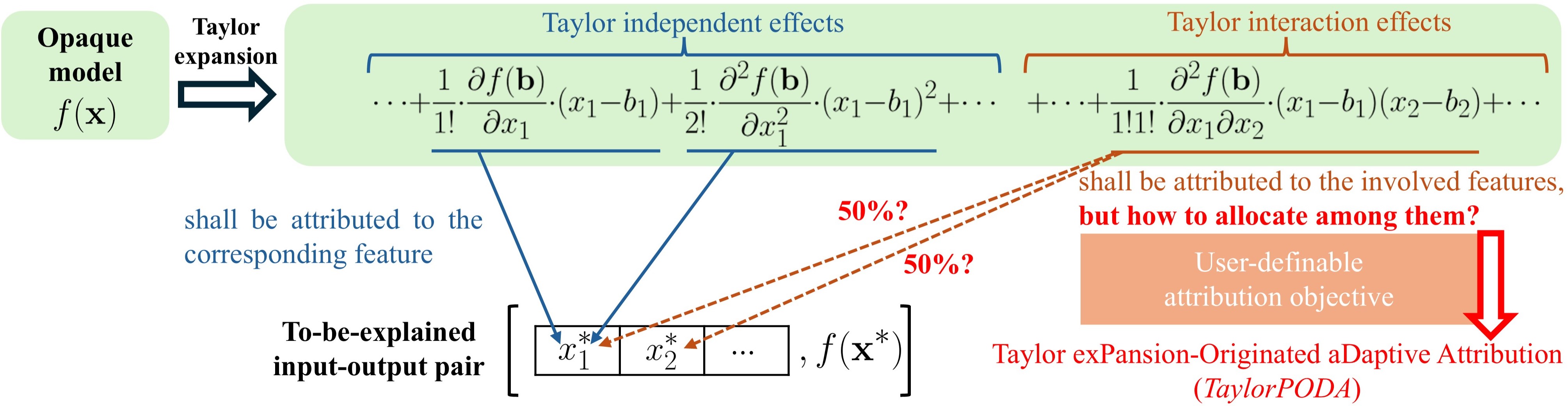}}
    \caption{Illustration of the Taylor expansion framework for analyzing LA methods, highlighting the potential for more flexible attribution design. While independent effects can be allocated in a targeted manner, interaction effects call for more nuanced consideration, particularly in model-agnostic settings with limited prior knowledge. Many well-recognized methods still rely on fixed, predefined allocations, e.g., SHAP adopts a uniform distribution among the features involved.}
    \label{figure_taylor_expansion}
\end{figure}

\section{From Taylor expansion to principled and adaptive post-hoc model-agnostic LA}
\label{s_preliminary}

In this section, we first describe how the Taylor expansion framework provides a principled perspective for analyzing post-hoc model-agnostic LA methods. Under this framework, we propose a set of postulates (\textit{precision}, \textit{federation}, and \textit{zero-discrepancy}) to regulate a theoretically grounded attribution mechanism. Based on these postulates, we analyze several existing post-hoc model-agnostic LA methods and show that the Shapley value satisfies all the proposed postulates. Nevertheless, previous studies have revealed limitations of the Shapley value in achieving an optimal explanation outcome due to its fixed and uniform attribution mechanism over marginal outputs. Although a more flexible variant, WeightedSHAP, improves empirical performance under particular evaluation preferences, it fails to preserve the proposed postulates. This reveals an important challenge in developing a post-hoc model-agnostic LA method that simultaneously satisfies all the proposed postulates while remaining compatible with diverse user needs.

\subsection{Taylor expansion framework for LA}
\label{ss_la_taylor_framework}
Let $\mathcal{X} \subseteq \mathbb{R}^n$ denote the input space with $n \in \mathbb{N}$, and let $\mathcal{Y} \subseteq \mathbb{R}$ denote the output space. A model $f: \mathcal{X} \rightarrow \mathcal{Y}$ maps an input point $\mathbf{x}$ to an output $f(\mathbf{x})$, where $\mathbf{x}=(x_1,\dots,x_n)$ and $x_i$ denotes the $i$-th feature of $\mathbf{x}$. For a baseline point $\mathbf{b}=(b_1,\dots,b_n)\in\mathbb{R}^n$, the $K$-order Taylor expansion of $f(\mathbf{x})$ around $\mathbf{b}$ is given by the following expression, assuming that $f$ is differentiable up to order $K$ in a neighborhood of $\mathbf{b}$ containing $\mathbf{x}$:
%
%
%
\begin{equation}
\label{eqn_taylor_fx}
\begin{aligned}
f(\mathbf{x})&=f(\mathbf{b})+\sum_{i=1}^{n}\frac{1}{1!}\cdot\frac{\partial f(\mathbf{b})}{\partial x_i}\cdot(x_i-b_i)+\sum_{i=1}^{n}\sum_{j=1}^{n}\frac{1}{2!}\cdot\frac{\partial^2 f(\mathbf{b})}{\partial x_i \partial x_j}\cdot(x_i-b_i)(x_j-b_j)+\dots+\epsilon_K\\
&=f(\mathbf{b})+\sum_{k=1}^{K}\sum_{\boldsymbol{\kappa}\in O_k}\underbrace{\frac{1}{(\kappa_1+\cdots+\kappa_n)!}\binom{\kappa_1+\cdots+\kappa_n}{\kappa_1,\ \dots,\ \kappa_n} \cdot \frac{\partial^{|\boldsymbol{\kappa}|}f(\mathbf{b})}{\partial x_1^{\kappa_1}\cdots\partial x_n^{\kappa_n}} \cdot \prod_{i=1}^{n}(x_i-b_i)^{\kappa_i}}_{I(\boldsymbol{\kappa})}+\epsilon_K,\\
\end{aligned}
\end{equation}
where $\boldsymbol{\kappa}=(\kappa_1,\dots,\kappa_n)\in O_k$ denotes a multi-index of order $k$, with $|\boldsymbol{\kappa}|=\sum_{i=1}^{n}\kappa_i=k$, $\mathcal{O}_k=\{\kappa\in\mathbb{N}_0^n:|\kappa|=k\}$, and $\epsilon_K$ denotes the approximation error induced by truncating the expansion at order $K$.
%
Within this additive decomposition, all Taylor terms 
$I(\boldsymbol{\kappa},\mathbf{b},\mathbf{x})$ (abbreviated as $I(\boldsymbol{\kappa})$ for simplicity) can be categorized into two types: \textit{Taylor interaction effects}, which arise from the joint involvement of multiple features, and \textit{Taylor independent effects}, which are associated with individual features:
\begin{definition}[Taylor interaction effect]
    \label{def_taylor_interaction}
    Taylor interaction effects are the additive terms in Equation~(\ref{eqn_taylor_fx}) given by $I(\boldsymbol{\kappa})$ for all $\boldsymbol{\kappa}\in \Omega_S$ with $|S|>1$, where $N=\{1,\dots,n\}$, $S\subseteq N$, and $\Omega_S=\{(\kappa_1,\dots,\kappa_n) \mid \text{ for all } i\in S,\ \kappa_i>0, \text{ and for all }i\notin S,\ \kappa_i=0\}.$
\end{definition}
\begin{definition}[Taylor independent effect]
    \label{def_taylor_independent}
    Taylor independent effects are the additive terms in Equation~(\ref{eqn_taylor_fx}) associated with individual features. Specifically, for $i\in N$, the Taylor independent effect corresponding to feature $i$ is defined as
    \begin{equation}
    \label{eqn_taylor_independent}
    \phi(\boldsymbol{\kappa}) := I(\boldsymbol{\kappa}),
    \text{ for all } \boldsymbol{\kappa}\in \Omega_{\{i\}}.
    \end{equation}
\end{definition}

Using Definitions \ref{def_taylor_interaction} and \ref{def_taylor_independent}, $f(\mathbf{x})$ can be rewritten as (ignoring the truncation error $\epsilon_K$): 
\begin{equation}
\label{eqn_taylor_fx_indint}
\begin{aligned}
f(\mathbf{x})=f(\mathbf{b}) + \sum_{i\in N}\sum_{\boldsymbol{\kappa}\in\Omega_{\{i\}}}\phi(\boldsymbol{\kappa}) + \sum_{\substack{S\subseteq N,|S|>1}}\sum_{\boldsymbol{\kappa}\in\Omega_{S}}I(\boldsymbol{\kappa}).
\end{aligned}
\end{equation}

Within the Taylor expansion framework involving Taylor independent effects and Taylor interaction effects, LA methods can be generalized in the following form:
\begin{definition}[Local attribution]
    \label{def_local_attribution}
    A local attribution method yields a vector of scores $\mathbf{a} = (a_1, \dots, a_n)$ with $a_i \in \mathbb{R}$ for $i\in N$, where $a_i$ quantifies the contribution of $x_i$ through a linear reallocation of the Taylor independent effects and the Taylor interaction effects:
    \begin{equation}
    \label{eqn_local_attribution}
    \begin{aligned}
    a_i = \sum_{j\in N}\sum_{\boldsymbol{\kappa}\in\Omega_{\{j\}}}\tau_{i,\boldsymbol{\kappa}}\phi(\boldsymbol{\kappa}) + \sum_{\substack{S\subseteq N,|S|>1}}\sum_{\boldsymbol{\kappa}\in\Omega_{S}}\zeta_{i,\boldsymbol{\kappa}}I(\boldsymbol{\kappa}),
    \end{aligned}
    \end{equation}
    where $\tau_{i,\boldsymbol{\kappa}},\zeta_{i,\boldsymbol{\kappa}}\in\mathbb{R}$. 
\end{definition}
%


\subsection{Three fundamental postulates}

Within the Taylor expansion framework discussed above, rigor in an LA method naturally entails attributing to each feature all and only the Taylor terms relevant to it. Accordingly, we introduce the following postulates to further regulate the LA defined in Equation~(\ref{eqn_local_attribution}) in a principled manner.
%
\begin{postulate}
    \label{postulate_precision}
    \textbf{\textit{Precision}}. The Taylor independent effect of the $i$-th feature shall be entirely attributed to the $i$-th feature, while it shall not be attributed to any other feature:
    \begin{equation}
    \label{eqn_presicion}
    \begin{aligned}
        \tau_{i,\kappa}
        =
        \begin{cases}
        1, & i=j,\\
        0, & i\neq j,
        \end{cases}
        \qquad
        \text{for }\kappa\in\Omega_{\{j\}}.
    \end{aligned}
    \end{equation}
\end{postulate}
\begin{postulate}
    \label{postulate_federation}
    \textbf{\textit{Federation}}. The Taylor interaction effect of the features in $S$ shall only be attributed to the features in $S$:
    \begin{equation}
    \label{eqn_federation}
            \zeta_{i,\boldsymbol{\kappa}} = 0, 
            \text{ for all } i \notin S, \ \boldsymbol{\kappa} \in \Omega_S.
    \end{equation}
\end{postulate}
\begin{postulate}
    \label{postulate_zero_discrepancy}
    \textbf{\textit{Zero-discrepancy}}. There should be neither redundancy nor deficiency in the attribution results regarding the allocation of the exact model output $f(\mathbf{x})$ to individual features. Equivalently, the value of discrepancy, denoted by $d$, shall equal zero:
    \begin{equation}
    \label{eqn_zero_discrepancy}
    \begin{aligned}
        d := f(\mathbf{b}) + \sum_{i\in N} a_i - f(\mathbf{x}) = 0.
    \end{aligned}
    \end{equation}
    This \textit{zero-discrepancy} postulate aligns with the design principle of \textit{local accuracy}, a core property of SHAP~\citep{lundberg2017unified}. Notably, \textit{local accuracy} is derived from the \textit{efficiency} axiom of the Shapley value~\citep{shapley1953value}, under which “the value represents a distribution of the full yield of the game.”
\end{postulate}

Among the proposed postulates, \textit{precision} and \textit{federation} regulate $a_i$ by attributing all Taylor terms involving the $i$-th feature while excluding unrelated terms. \textit{Zero-discrepancy} ensures that the base value plus the sum of the attributions exactly matches the model output, thereby fully allocating the Taylor terms among the features. To illustrate the necessity of these postulates, Table~\ref{table_postulate_conseq} presents examples within the Taylor framework (under a three-feature setting, $N = \{1, 2, 3\}$). These cases show how omitting any single postulate may lead to non-rigorous results.
\renewcommand{\arraystretch}{1.2}
\begin{table}[hptb]
\caption{Illustrative examples of the effects of omitting individual postulates.}
\label{table_postulate_conseq}
\begin{center}
\footnotesize
\begin{tabular}{c c l}
\toprule
\textbf{Absent postulate} & \textbf{Example Taylor term} & \multicolumn{1}{c}{\textbf{Potential effects}} \\
\midrule
\textit{Precision} & $\frac{1}{5!}\cdot\frac{\partial^5f(\mathbf{b})}{\partial x_2^5}\cdot(x_2-b_2)^5$ & \makecell[l]{\textbf{The} Taylor term can be attributed to $x_1$ and \\included in $a_1$, rather than being fully attributed\\to $x_2$ and included in $a_2$. }\\
\textit{Federation} & $\frac{1}{2!}\cdot\frac{\partial^2f(\mathbf{b})}{\partial x_1\partial x_2}\cdot(x_1-b_1)\cdot(x_2-b_2)$ & \makecell[l]{\textbf{The} Taylor term can be attributed to $x_3$ and\\included in $a_3$. } \\
\makecell{\textit{Zero}\\\textit{-discrepancy}} & $\frac{1}{3!}\cdot\frac{\partial^3f(\mathbf{b})}{\partial x_1\partial x_2\partial x_3}\cdot(x_1-b_1)\cdot(x_2-b_2)\cdot(x_3-b_3)$ & \makecell[l]{\textbf{The} Taylor term can be incompletely allocated\\(e.g., only 90\% distributed among \( a_1, a_2, a_3 \)),\\which leads to \( f(\mathbf{b}) + a_1 + a_2 + a_3 \neq f(\mathbf{x}) \).
}\\
\bottomrule
\end{tabular}
\end{center}
\end{table}

\subsection{Existing post-hoc model-agnostic LA methods and their limitations}
\label{ss_existing}

Since masked outputs of $f$ are widely used in existing LA methods, we formally specify their definition here.
\begin{definition}[Masked output]
    \label{def_masked_output}
    Given an input $\mathbf{x}$, the masked output of $f$ with respect to a feature subset $S\subseteq N$ is defined as
    \begin{equation}
    \label{eqn_masked_output}
    \begin{aligned}
        f_S(\mathbf{x})
        :=
        \mathbb{E}\left[
        f(\mathbf{X})
        \mid
        X_i = x_i,\text{ for all } i\in S
        \right],
    \end{aligned}
    \end{equation}
    which estimates the model output when the features in $S$ are fixed to their observed values in $\mathbf{x}$, while the remaining features are marginalized out according to the data distribution.
\end{definition}

\cite{deng2024unifying} explored fourteen post-hoc LA methods by leveraging the Taylor expansion framework outlined in Equation (\ref{eqn_local_attribution}). Among them, four methods (Occlusion-1, Occlusion-patch, Prediction Diff, and Shapley value) are model-agnostic, while others require model-specific internal information (e.g., model gradients for Grad-CAM). Moreover, because Occlusion-patch and Prediction Diff can be regarded as variants of Occlusion-1 with similar mechanisms, we use Occlusion-1 (OCC-1) and the Shapley value as representative examples. In addition, we include and analyze two other widely adopted methods, LIME and WeightedSHAP, to support the subsequent comparative analysis.

\textbf{OCC-1: } 
\begin{equation}
\label{eqn_occ}
\begin{aligned}
\underset{(\text{OCC-1})}{a_i}
=f(\mathbf{x})-f_{N\backslash\{i\}}(\mathbf{x})
=\sum_{\boldsymbol{\kappa}\in\Omega_{\{i\}}}\phi(\boldsymbol{\kappa}) + \sum_{\substack{S\subseteq N\\|S|>1\\S\supseteq\{i\}}}\sum_{\boldsymbol{\kappa}\in\Omega_{S}}I(\boldsymbol{\kappa}).
\end{aligned}
\end{equation}

\textbf{Shapley value: }
\begin{equation}
\label{eqn_shapley_value}
\begin{aligned}
\underset{(\text{Shapley value})}{a_i}
=\sum_{S\subseteq N\backslash\{i\}}p(S)\cdot\left[f_{S\cup\{i\}}(\mathbf{x})-f_S(\mathbf{x})\right]
=\sum_{\boldsymbol{\kappa}\in\Omega_{\{i\}}}\phi(\boldsymbol{\kappa})+\sum_{\substack{S\subseteq N\\|S|>1\\S\supseteq\{i\}}}\sum_{\boldsymbol{\kappa}\in \Omega_S}\frac{1}{|S|}I(\boldsymbol{\kappa}),
\end{aligned}
\end{equation}
where $p(S)=|S|!(|N|-1-|S|)!/|N|!$, which is determined by the cardinality of $S$ and $N$. (In the remainder of this paper, “Shapley value” is used as a generic term for LA methods derived from the cooperative game-theoretic Shapley value, including SHAP unless otherwise defined.)

\textbf{WeightedSHAP: }
\begin{equation}
\label{eqn_weightedshap}
\begin{aligned}
&\underset{(\text{WeightedSHAP})}{a_i}
=\sum_{S\subseteq N\backslash\{i\}}w_S\left[f_{S\cup\{i\}}(\mathbf{x})-f_S(\mathbf{x})\right]
=\sum_{S\subseteq N\backslash\{i\}}w_S\left[\sum_{\boldsymbol{\kappa}\in\Omega_{\{i\}}}\phi(\boldsymbol{\kappa})+\sum_{\substack{T \subseteq S \cup \{i\}\\T\supseteq \{i\}}}\sum_{\boldsymbol{\kappa}\in \Omega_T}I(\boldsymbol{\kappa})\right],\\
\end{aligned}
\end{equation}
where $w_S$ denotes an adaptive weight assigned to the marginal contribution associated with subset $S$. WeightedSHAP allows these weights to vary within a flexible family of valid weight vectors, with the specific selection procedure detailed in \cite{kwonWeightedSHAPAnalyzingImproving2022}.

\textbf{LIME: }
Instead of quantifying each feature’s contribution by calculating and aggregating masked outputs, LIME fits a local surrogate model $g: \mathcal{X} \rightarrow \mathcal{Y}$. In practice, $g$ is typically defined as a linear model to approximate $f(\mathbf{x})$:
\begin{equation}
\label{eqn_lime}
\begin{aligned}
&\underset{(\text{LIME})}{g(\mathbf{x})}
=\sum_{i\in N}\eta_i\cdot x_i\approx f(\mathbf{x}),\\
\end{aligned}
\end{equation}
where the coefficient $\eta_i$ measures the contribution of the
$i$-th feature, and $\boldsymbol{\eta} = (\eta_1, \dots, \eta_n)$ is given by
\begin{equation}
\label{eqn_lime_argmin}
\begin{aligned}
\boldsymbol{\eta}^*=\arg\min_{\boldsymbol{\eta}\in\mathbb{R}^n}\mathcal{L}(f,g_{\boldsymbol{\eta}},\iota_{\mathbf{x}})+\theta(g_{\boldsymbol{\eta}}),\\
\end{aligned}
\end{equation}
with $\iota_\mathbf{x}$ defining the weighting for the loss function $\mathcal{L}$, and $\theta(g_{\boldsymbol{\eta}})$ regulating the complexity of $g_{\boldsymbol{\eta}}$. 



\textbf{Important limitations remain in existing post-hoc model-agnostic LA methods.} Among these methods, OCC-1 and WeightedSHAP violate \textit{zero-discrepancy}, leading to redundancy or deficiency between the model output and the resulting attributions. In addition, WeightedSHAP violates \textit{precision}, meaning that Taylor independent effects are not attributed exclusively and exactly to their corresponding features. LIME falls outside the scope of these Taylor expansion-based postulates, since its attribution results are derived from a surrogate model rather than from reallocating the effects contributing to the original model output. Furthermore, its explainability may be questioned because the surrogate model fails to capture the nonlinearity of the original model~\citep{salih2025perspective}. In contrast, the Shapley value satisfies all the proposed postulates and appears to be the only existing post-hoc model-agnostic LA method with this set of theoretical guarantees. Nevertheless, the Shapley value does not consistently achieve optimal performance with respect to the “user-defined utility”~\citep{kwonWeightedSHAPAnalyzingImproving2022}. Such utilities quantify the “goodness” of attribution according to the user’s downstream purpose. Specifically, this limitation arises from the fixed and uniform attribution mechanism adopted over marginal contributions. Accordingly, \cite{kwonWeightedSHAPAnalyzingImproving2022} proposed WeightedSHAP, which learns a weight vector to optimize attribution for a user-defined utility. However, WeightedSHAP fails to satisfy \textit{zero-discrepancy} and \textit{precision}, weakening the theoretical grounding of its attribution process. (A detailed analysis of how each method satisfies or violates these postulates is provided in Section~\ref{s_theoretical_analysis}.)

\subsection{The challenge: bridging postulates and practical utilities}

The picture thus becomes clearer. Satisfying all the postulates ensures a theoretically grounded attribution mechanism, but does not necessarily guarantee superior explainability “goodness”. This is because there is often no universally optimal explanation outcome in post-hoc model-agnostic scenarios. Instead, the usefulness of LA results is inherently tied to how effectively they support downstream application needs. Some existing approaches that employ adaptable attribution of Taylor terms may improve empirical behavior in certain applications, but can simultaneously violate the underlying postulates, thereby weakening the theoretical consistency of the attribution process. This reveals a fundamental tension between theoretical rigor and practical utilities in existing methods. 

In this regard, we seek to develop a new method that
\begin{itemize}
    \item satisfies all the postulates, and
    \item retains the flexibility to incorporate user-defined utilities.
\end{itemize}

\section{Proposed method}
\label{s_proposed_method}



We propose \textbf{TaylorPODA}, which attributes the output $f(\mathbf{x})$ to the $i$-th feature as follows: 
\begin{equation}
\label{eqn_taylorpoda}
\begin{aligned}
    \underset{(\text{TaylorPODA})}{a_i}
    =f(\mathbf{x})-f_{N\backslash\{i\}}(\mathbf{x})-\sum_{\substack{S\subseteq N\\|S|>1\\S\supseteq\{i\}}}(1-\xi_{i, S})H(S),
\end{aligned}
\end{equation}
where the Harsanyi dividend $H(S) = \sum_{T \subseteq S} (-1)^{|T|-|S|} f_T(\mathbf{x})$, originally proposed by \cite{harsanyi1982simplified} for game-theoretic analysis, is here applied as an operator over masked model outputs, and the coefficients $\xi_{i,S}\in [0,1]$ are tunable weights introduced to enable further adaptation based on a user-defined utility. Specifically, without loss of generality, we formulate the optimization problem as a maximization problem:
\begin{equation}
\label{eqn_taylorpoda_obj}
\begin{aligned}
\boldsymbol{\xi}^*
=
&\operatorname*{argmax}_{\boldsymbol{\xi}\in\mathcal{Z}}
u(\boldsymbol{\xi};f,\mathbf{x}),
\\
\text{s.t.}\quad
&
\sum_{i\in S} \xi_{i,S}
=
1,
\quad
\text{for all } S\subseteq N,\ |S|>1,
\end{aligned}
\end{equation}
where $\mathcal{Z}$ denotes the set of all valid coefficient vectors, and $u(\boldsymbol{\xi};f,\mathbf{x})$ denotes a user-defined utility function that defines the optimization target. Minimization objectives can be equivalently incorporated by maximizing their negative values.

Notably, the proposed formulation of TaylorPODA is designed to support the property of \textit{adaptation}, allowing the attribution process to remain compatible with diverse practical settings through flexible allocation of Taylor interaction effects. 
%
%
\begin{property} 
\label{property_adaptation} 
    \textbf{\textit{Adaptation}}. For the $i$-th feature, the proportion of each Taylor interaction effect allocated to this feature shall be tunable. Specifically, the attribution mechanism allows $\zeta_{i, \boldsymbol{\kappa}} \in \left[ 0, 1 \right]$ for all $\boldsymbol{\kappa} \in \Omega_S$ with $S\subseteq N, |S| > 1,$ and $i \in S$.
\end{property} 
This property enables the attribution process to flexibly allocate Taylor interaction effects among the features in $S$. It is motivated by the findings of \cite{kwonWeightedSHAPAnalyzingImproving2022}, which show that the fixed and predefined attribution mechanism adopted by Shapley-value-based methods can be suboptimal under user-defined utilities. As demonstrated by \cite{kwonWeightedSHAPAnalyzingImproving2022}, the Shapley value “assigns uniform weights to marginal contributions in computing the attribution score ... this can lead to attribution mistakes when different marginal contributions have different signal and noise.” Within the Taylor expansion framework, these “uniform weights” correspond to the uniform allocation of Taylor interaction effects among the involved features (see Equation~(\ref{eqn_shapley_value})). 


By relaxing this fixed and predefined mechanism, the proposed \textit{adaptation} property introduces flexibility into the attribution process. This flexibility enables the attribution process to be optimized with respect to user-defined utilities. Furthermore, beyond the learnable attribution weights introduced in WeightedSHAP, TaylorPODA realizes such flexibility in a more principled manner. Specifically, \textit{adaptation} is introduced solely through the allocation of Taylor interaction effects, while the attribution of Taylor independent effects remains unaffected. This preserves the direct correspondence between independent effects and their associated features. 

A detailed analysis of the \textit{adaptation} property is provided in Section~\ref{s_theoretical_analysis}, where we show that existing post-hoc model-agnostic LA methods do not fully satisfy this property, whereas TaylorPODA is designed to achieve it. 

\section{Theoretical analysis: postulate satisfaction}
\label{s_theoretical_analysis}

\textbf{TaylorPODA satisfies Postulate~\ref{postulate_precision} (\textit{precision}), Postulate~\ref{postulate_federation} (\textit{federation}), and Postulate \ref{postulate_zero_discrepancy} (\textit{zero-discrepancy}), and supports Property~\ref{property_adaptation} (\textit{adaptation}}).

\textbf{Proof: }

By Theorem 2 in the work of \cite{deng2024unifying}, we have:
\begin{equation}
\label{eqn_taylor_ff}
    \begin{aligned}
        f(\mathbf{x})-f_{N\backslash\{i\}}(\mathbf{x})
        =\sum_{\boldsymbol{\kappa}\in\Omega_{\{i\}}}\phi(\boldsymbol{\kappa})+\sum_{\substack{S\subseteq N\\|S|>1\\S\supseteq\{i\}}}\sum_{\boldsymbol{\kappa}\in \Omega_S}I(\boldsymbol{\kappa}).
    \end{aligned}
\end{equation}
Also, by Theorem 1 in the work of \cite{deng2024unifying}, we have
\begin{equation}
\label{eqn_taylor_harsanyi}
    \begin{aligned}
        H(S) = \sum_{\boldsymbol{\kappa}\in \Omega_S}I(\boldsymbol{\kappa}), \text{ for all } S\subseteq N, |S|>1.\\
    \end{aligned}
\end{equation}
Substituting Equations (\ref{eqn_taylor_ff}) and (\ref{eqn_taylor_harsanyi}) into Equation (\ref{eqn_taylorpoda}), we get
\begin{equation}
\label{eqn_taylor_term}
    \begin{aligned}
        \underset{(\text{TaylorPODA})}{a_i}
        &=f(\mathbf{x})-f_{N\backslash\{i\}}(\mathbf{x})-\sum_{\substack{S\subseteq N\\|S|>1\\S\supseteq\{i\}}}(1-\xi_{i,S})H(S)\\
        &=\sum_{\boldsymbol{\kappa}\in\Omega_{\{i\}}}\phi(\boldsymbol{\kappa})+\sum_{\substack{S\subseteq N\\|S|>1\\S\supseteq\{i\}}}\sum_{\boldsymbol{\kappa}\in \Omega_S}I(\boldsymbol{\kappa})-\sum_{\substack{S\subseteq N\\|S|>1\\S\supseteq\{i\}}}(1-\xi_{i,S})H(S)\\
        &=\sum_{\boldsymbol{\kappa}\in\Omega_{\{i\}}}\phi(\boldsymbol{\kappa})+\sum_{\substack{S\subseteq N\\|S|>1\\S\supseteq\{i\}}}H(S)-\sum_{\substack{S\subseteq N\\|S|>1\\S\supseteq\{i\}}}(1-\xi_{i,S})H(S)\\
        &=\sum_{\boldsymbol{\kappa}\in\Omega_{\{i\}}}\phi(\boldsymbol{\kappa})+\sum_{\substack{S\subseteq N\\|S|>1\\S\supseteq\{i\}}}\sum_{\boldsymbol{\kappa}\in \Omega_S}\xi_{i,S}I(\boldsymbol{\kappa}).\\
    \end{aligned}
\end{equation}

By setting $\tau_{i,\kappa}=1$ for $\kappa\in\Omega_{\{i\}}$, $\tau_{i,\kappa}=0$ for $\kappa\in\Omega_{\{j\}}$ with $j\neq i$, and $\zeta_{i,\kappa}:=\xi_{i,S}$ for $\kappa\in\Omega_S$, Equation (\ref{eqn_local_attribution}) becomes equivalent to Equation (\ref{eqn_taylor_term}). Therefore, when calculating $\underset{(\text{TaylorPODA})}{a_i}$, only the Taylor independent effects associated with the $i$-th feature, i.e., $\sum_{\boldsymbol{\kappa}\in\Omega_{\{i\}}}\phi(\boldsymbol{\kappa})$, and no Taylor independent effects associated with any feature $j\neq i$ are involved. \textbf{Thus, TaylorPODA satisfies Postulate~\ref{postulate_precision}.} Similarly, according to Equations (\ref{eqn_occ}) and (\ref{eqn_shapley_value}), OCC-1 and SHAP satisfy Postulate~\ref{postulate_precision}. As demonstrated in Equation (\ref{eqn_weightedshap}), WeightedSHAP attributes Taylor independent effects with a weighting factor $w_S$, thereby violating Postulate~\ref{postulate_precision}. 

As indicated in Equation (\ref{eqn_taylor_term}), Taylor interaction effects will be attributed to the $i$-th feature if and only if $S\subseteq N\ $ with $\ S\supseteq\{i\} $ and $ |S|>1$. \textbf{Thus, TaylorPODA satisfies Postulate~\ref{postulate_federation}.} Similarly, Equations (\ref{eqn_occ}), (\ref{eqn_shapley_value}), and (\ref{eqn_weightedshap}) show that OCC-1, SHAP, and WeightedSHAP satisfy Postulate~\ref{postulate_federation}.

For \textit{zero-discrepancy}, we have
\begin{equation}
\label{eqn_taylorpoda_integrity}
    \begin{aligned}
        & f(\mathbf{b}) + \sum_{i\in N} \underset{(\text{TaylorPODA})}{a_i}\\
        = & f(\mathbf{b}) + \sum_{i\in N} \left[
        \sum_{\boldsymbol{\kappa}\in\Omega_{\{i\}}}\phi(\boldsymbol{\kappa})+\sum_{\substack{S\subseteq N\\|S|>1\\S\supseteq\{i\}}}\xi_{i,S}\sum_{\boldsymbol{\kappa}\in \Omega_S}I(\boldsymbol{\kappa})
        \right]\\
        = & f(\mathbf{b}) + 
        \sum_{i\in N}\sum_{\boldsymbol{\kappa}\in\Omega_{\{i\}}}\phi(\boldsymbol{\kappa})+\sum_{i\in N}\sum_{\substack{S\subseteq N\\|S|>1\\S\supseteq\{i\}}}\xi_{i,S}\sum_{\boldsymbol{\kappa}\in \Omega_S}I(\boldsymbol{\kappa}).
        \\
    \end{aligned}
\end{equation}
Given that $\sum_{i\in S}\xi_{i,S}=1$ for every $S\subseteq N$ with $|S|>1$, Equation (\ref{eqn_taylorpoda_integrity}) can be further transformed into:
\begin{equation}
\label{eqn_taylorpoda_integrity2}
    \begin{aligned}
        f(\mathbf{b}) + \sum_{i\in N} \underset{(\text{TaylorPODA})}{a_i}
        =
        f(\mathbf b)
        +\sum_{i\in N}\sum_{\kappa\in\Omega_{\{i\}}}\phi(\kappa)
        +\sum_{\substack{S\subseteq N\\|S|>1}}
        \sum_{\kappa\in\Omega_S}I(\kappa) 
        = f(\mathbf x).
    \end{aligned}
\end{equation}
\textbf{Thus, TaylorPODA satisfies Postulate \ref{postulate_zero_discrepancy}.} Similarly, as given in Equation (\ref{eqn_occ}) for OCC-1, $\xi_{i,S}=1$ for every $i\in S$, and therefore $\sum_{i\in S}\xi_{i,S}=|S|>1.$ Thus, OCC-1 violates zero-discrepancy. Likewise, as given in Equation (\ref{eqn_shapley_value}) for the Shapley value, $\xi_{i,S}=1/|S|$ for every $i\in S$, and hence $\sum_{i\in S}\xi_{i,S}=1$. Thus, the Shapley value satisfies zero-discrepancy. However, it is not guaranteed that $\sum_{i\in S}\xi_{i,S}=1$ in WeightedSHAP, since the weighting factors are not constrained to sum to one. Thus, the Shapley value satisfies Postulate \ref{postulate_zero_discrepancy}, whereas OCC-1 and WeightedSHAP violate Postulate \ref{postulate_zero_discrepancy}.

Moreover, as $\xi_{i,S}$ is an adaptive weight, the proportion of each Taylor interaction effect allocated to the $i$-th feature is adjustable rather than fixed. \textbf{Thus, TaylorPODA introduces Property \ref{property_adaptation}.} Similarly, Equations (\ref{eqn_occ}), (\ref{eqn_shapley_value}), and (\ref{eqn_weightedshap}) show that WeightedSHAP partially satisfies Property \ref{property_adaptation} since it does not impose the constraint
$\zeta_{i,\kappa}\in[0,1]$, whereas OCC-1 and SHAP violate Property \ref{property_adaptation}.

LIME introduces an external surrogate model $g(\mathbf{x})$ to approximate the original model, as shown by Equations (\ref{eqn_lime}) and (\ref{eqn_lime_argmin}). Consequently, LIME falls outside the scope of the postulate and property system in this work.

The satisfaction of the proposed postulates and properties is summarized in Table~\ref{table_postulate_satisfaction}.
\begin{table}[thb]
\begin{center}
\caption{Comparison of postulate and property satisfaction across existing LA methods. TaylorPODA and SHAP are the only methods satisfying all postulates, whereas only TaylorPODA additionally satisfies the \textbf{adaptation} property.}
\label{table_postulate_satisfaction}
\small
\begin{tabular}{ccccc}
\toprule
\multirow{2}{*}{\textbf{Method}}
& \multicolumn{3}{c}{\textbf{Postulates}}
& \multicolumn{1}{c}{\textbf{Property}} \\
\cmidrule(lr){2-4} 
& \textbf{\textit{Precision}}
& \textbf{\textit{Federation}}
& \textbf{\textit{Zero-discrepancy}}
& \textbf{\textit{Adaptation}} \\
\midrule
OCC-1   & $\checkmark$ & $\checkmark$ & $\times$     & $\times$ \\
LIME          & --           & --           & --           & --       \\
SHAP          & $\checkmark$ & $\checkmark$ & $\checkmark$ & $\times$ \\
WeightedSHAP  & $\times$     & $\checkmark$ & $\times$     & Partial \\
TaylorPODA    & $\checkmark$ & $\checkmark$ & $\checkmark$ & $\checkmark$ \\
\bottomrule
\end{tabular}
\end{center}
\end{table}


\textbf{Furthermore, TaylorPODA remains principled and capable of explaining non-differentiable models. It satisfies the analogous forms of Postulates~\ref{postulate_precision}, \ref{postulate_federation}, \ref{postulate_zero_discrepancy}, and Property~\ref{property_adaptation} by attributing Harsanyi dividends instead of Taylor terms.} Although TaylorPODA is motivated by the Taylor expansion framework, which assumes model differentiability, the formulation in Equation (\ref{eqn_taylorpoda}) applies equally to non-differentiable models while retaining a principled attribution mechanism. This is because TaylorPODA also provides a clear and rigorous attribution mechanism by allocating Harsanyi dividends in a manner analogous to the allocation of Taylor terms. Essentially, Equation (\ref{eqn_taylorpoda}) is equivalent to
\begin{equation}
\label{eqn_taylorpoda_har_equiv}
\begin{aligned}
    \underset{(\text{TaylorPODA})}{a_i}
    &=f(\mathbf{x})-f_{N\backslash\{i\}}(\mathbf{x})-\sum_{\substack{S\subseteq N\\|S|>1\\S\supseteq\{i\}}}(1-\xi_{i, S})H(S)
    =H(\{i\}) + \sum_{\substack{S\subseteq N\\|S|>1\\S\supseteq\{i\}}}\xi_{i, S}H(S),\\
\end{aligned}
\end{equation}
which does not depend on model differentiability. When switching to this Harsanyi-dividend formulation, the analogous forms of Postulates~\ref{postulate_precision}, \ref{postulate_federation}, \ref{postulate_zero_discrepancy}, and Property~\ref{property_adaptation} can be defined directly, and they remain satisfied without reference to Taylor terms. Following the same rationale, only the Harsanyi dividends involving feature $i$ are explicitly and exhaustively assigned to $a_i$. In cooperative game theory, each Harsanyi dividend quantifies the additional joint effect arising purely from the act of cooperation itself; e.g., for a feature set (coalition) $S$, $H(S)$ represents the intrinsic synergistic contribution generated only when the features (players) in $S$ act together. 

\textit{Proof}: The proof of Equation (\ref{eqn_taylorpoda_har_equiv}), together with the verification that TaylorPODA satisfies the analogous Harsanyi-dividend forms of Postulates~\ref{postulate_precision}, \ref{postulate_federation}, \ref{postulate_zero_discrepancy}, and Property~\ref{property_adaptation}, is provided in Appendix \ref{apdx_harsanyi_analog}. \hfill $\blacksquare$

\renewcommand{\arraystretch}{1.2}
\begin{table}[hpbt]
\caption{Adaptation performance under different user-defined utility objectives on classification tasks. Values denote differences from the Shapley value baseline. ($\uparrow$ and $\downarrow$ indicate changes toward better and worse performance, respectively.)}
\begin{center}
\footnotesize
\begin{tabular}{cc c llll}
\toprule
\makecell{\textbf{Model}} & \makecell{\textbf{Data}} & \makecell{\textbf{Method}} & \makecell{\textbf{Inclusion AUP}\\\tiny{($\times10^{-2}$)}} & \makecell{\textbf{Exclusion AUP}\\\tiny{($\times10^{-2}$)}} & \makecell{\textbf{Inclusion AUC}\\\tiny{($\times10^{-2}$)}} & \makecell{\textbf{Exclusion AUC}\\\tiny{($\times10^{-2}$)}} \\
\midrule

\multirow{12}{*}{\rotatebox{90}{\makecell{MLP \tiny{(\texttt{tanh} \& \texttt{logistic})}}}}
& \multirow{4}{*}{\rotatebox{90}{Cancer}}
  & OCC-1 
& $\downarrow$61.0 {\tiny ($\downarrow$76.0, $\downarrow$47.8)} & $\downarrow$52.7 {\tiny ($\downarrow$73.1, $\downarrow$32.0)}
& $\downarrow$3.2 {\tiny ($\downarrow$5.4, $\downarrow$1.2)} & $\uparrow$0.3 {\tiny ($\downarrow$3.4, $\uparrow$4.6)} \\
& & LIME
& $\uparrow$0.1 {\tiny ($\downarrow$3.1, $\uparrow$3.1)} & $\downarrow$35.5 {\tiny ($\downarrow$44.1, $\downarrow$26.7)}
& $\uparrow$0.2 {\tiny ($\downarrow$0.5, $\uparrow$0.9)} & $\downarrow$2.6 {\tiny ($\downarrow$4.8, $\uparrow$0.0)} \\
& & WeightedSHAP
& $\uparrow$\textbf{7.6} {\tiny ($\uparrow$3.4, $\uparrow$11.9)} & $\uparrow$\textbf{33.9} {\tiny ($\uparrow$21.9, $\uparrow$45.9)}
& $\uparrow$\textbf{0.6} {\tiny ($\uparrow$0.0, $\uparrow$1.2)} & $\uparrow$\textbf{7.4} {\tiny ($\uparrow$4.1, $\uparrow$10.7)} \\
& & TaylorPODA
& $\uparrow$7.2 {\tiny ($\uparrow$4.2, $\uparrow$10.3)} & $\uparrow$24.2 {\tiny ($\uparrow$16.4, $\uparrow$31.2)}
& $\uparrow$0.4 {\tiny ($\uparrow$0.0, $\uparrow$1.0)} & $\uparrow$4.8 {\tiny ($\uparrow$2.7, $\uparrow$6.3)} \\
\cmidrule(lr){2-7}
& \multirow{4}{*}{\rotatebox{90}{Rice}} 
  & OCC-1 
& $\downarrow$27.3 {\tiny ($\downarrow$34.7, $\downarrow$19.8)} & $\downarrow$61.2 {\tiny ($\downarrow$71.2, $\downarrow$51.2)}
& $\downarrow$1.9 {\tiny ($\downarrow$3.6, $\downarrow$0.9)} & $\downarrow$4.6 {\tiny ($\downarrow$7.2, $\downarrow$1.9)} \\
& & LIME
& $\downarrow$2.2 {\tiny ($\downarrow$5.1, $\uparrow$0.4)} & $\downarrow$36.7 {\tiny ($\downarrow$42.2, $\downarrow$29.3)}
& $\uparrow$0.1 {\tiny ($\downarrow$1.2, $\uparrow$1.3)} & $\downarrow$2.9 {\tiny ($\downarrow$5.4, $\downarrow$0.4)} \\
& & WeightedSHAP
& $\uparrow$2.2 {\tiny ($\uparrow$0.8, $\uparrow$3.8)} & $\uparrow$6.9 {\tiny ($\uparrow$3.8, $\uparrow$10.7)}
& $\uparrow$0.6 {\tiny ($\uparrow$0.1, $\uparrow$1.1)} & $\uparrow$3.6 {\tiny ($\uparrow$2.1, $\uparrow$5.1)} \\
& & TaylorPODA
& $\uparrow$\textbf{4.2} {\tiny ($\uparrow$2.5, $\uparrow$6.5)} & $\uparrow$\textbf{8.9} {\tiny ($\uparrow$5.1, $\uparrow$13.9)}
& $\uparrow$\textbf{1.1} {\tiny ($\uparrow$0.4, $\uparrow$2.0)} & $\uparrow$\textbf{5.3} {\tiny ($\uparrow$3.5, $\uparrow$7.2)} \\
\cmidrule(lr){2-7}
& \multirow{4}{*}{\rotatebox{90}{Titanic}}
  & OCC-1 
& $\uparrow$4.2 {\tiny ($\uparrow$1.5, $\uparrow$7.6)} & $\uparrow$5.1 {\tiny ($\downarrow$8.0, $\uparrow$18.2)}
& $\downarrow$0.9 {\tiny ($\downarrow$3.1, $\uparrow$1.5)} & $\uparrow$9.9 {\tiny ($\uparrow$6.1, $\uparrow$14.1)} \\
& & LIME
& $\downarrow$10.6 {\tiny ($\downarrow$14.3, $\downarrow$7.4)} & $\downarrow$14.5 {\tiny ($\downarrow$20.0, $\downarrow$8.2)}
& $\downarrow$1.9 {\tiny ($\downarrow$3.4, $\uparrow$0.4)} & $\downarrow$6.9 {\tiny ($\downarrow$8.8, $\downarrow$4.4)} \\
& & WeightedSHAP
& $\uparrow$\textbf{9.8} {\tiny ($\uparrow$7.2, $\uparrow$12.6)} & $\uparrow$\textbf{33.9} {\tiny ($\uparrow$25.5, $\uparrow$43.8)}
& $\uparrow$\textbf{2.1} {\tiny ($\uparrow$1.0, $\uparrow$3.8)} & $\uparrow$\textbf{11.4} {\tiny ($\uparrow$8.1, $\uparrow$15.1)} \\
& & TaylorPODA
& $\uparrow$5.2 {\tiny ($\uparrow$3.9, $\uparrow$7.2)} & $\uparrow$16.0 {\tiny ($\uparrow$14.2, $\uparrow$17.7)}
& $\uparrow$1.7 {\tiny ($\uparrow$1.0, $\uparrow$2.7)} & $\uparrow$3.6 {\tiny ($\uparrow$2.0, $\uparrow$5.7)} \\

\cmidrule(lr){1-7}
\multirow{12}{*}{\rotatebox{90}{\makecell{MLP \tiny{(\texttt{ReLU})}}}}
& \multirow{4}{*}{\rotatebox{90}{Cancer}}
  & OCC-1 
& $\downarrow$97.6 {\tiny ($\downarrow$122.2, $\downarrow$75.3)} & $\downarrow$94.9 {\tiny ($\downarrow$116.5, $\downarrow$76.7)}
& $\downarrow$10.0 {\tiny ($\downarrow$13.6, $\downarrow$7.2)} & $\downarrow$9.4 {\tiny ($\downarrow$14.2, $\downarrow$6.0)} \\
& & LIME
& $\downarrow$14.6 {\tiny ($\downarrow$21.4, $\downarrow$10.7)} & $\downarrow$48.3 {\tiny ($\downarrow$59.7, $\downarrow$38.8)}
& $\downarrow$1.2 {\tiny ($\downarrow$2.5, $\downarrow$0.4)} & $\downarrow$5.6 {\tiny ($\downarrow$8.3, $\downarrow$3.7)} \\
& & WeightedSHAP
& $\uparrow$\textbf{4.2} {\tiny ($\uparrow$0.4, $\uparrow$11.0)} & $\uparrow$\textbf{12.9} {\tiny ($\uparrow$4.8, $\uparrow$22.6)}
& $\uparrow$0.3 {\tiny ($\uparrow$0.0, $\uparrow$1.1)} & $\uparrow$1.4 {\tiny ($\uparrow$0.2, $\uparrow$3.3)} \\
& & TaylorPODA
& $\uparrow$\textbf{4.2} {\tiny ($\uparrow$1.5, $\uparrow$8.8)} & $\uparrow$12.7 {\tiny ($\uparrow$6.3, $\uparrow$19.2)}
& $\uparrow$\textbf{0.8} {\tiny ($\uparrow$0.2, $\uparrow$1.8)} & $\uparrow$\textbf{1.7} {\tiny ($\uparrow$0.5, $\uparrow$2.8)} \\
\cmidrule(lr){2-7}
& \multirow{4}{*}{\rotatebox{90}{Rice}} 
  & OCC-1 
& $\downarrow$26.1 {\tiny ($\downarrow$33.9, $\downarrow$18.4)} & $\downarrow$67.3 {\tiny ($\downarrow$75.0, $\downarrow$58.3)}
& $\downarrow$4.3 {\tiny ($\downarrow$6.2, $\downarrow$2.4)} & $\downarrow$8.9 {\tiny ($\downarrow$11.4, $\downarrow$6.7)} \\
& & LIME
& $\downarrow$0.1 {\tiny ($\downarrow$2.0, $\uparrow$2.0)} & $\downarrow$14.7 {\tiny ($\downarrow$24.8, $\downarrow$7.4)}
& $\downarrow$0.4 {\tiny ($\downarrow$2.0, $\uparrow$1.7)} & $\downarrow$0.9 {\tiny ($\downarrow$3.4, $\uparrow$2.5)} \\
& & WeightedSHAP
& $\uparrow$2.9 {\tiny ($\uparrow$1.6, $\uparrow$4.7)} & $\uparrow$8.6 {\tiny ($\uparrow$5.6, $\uparrow$11.8)}
& $\uparrow$0.7 {\tiny ($\uparrow$0.1, $\uparrow$1.7)} & $\uparrow$3.1 {\tiny ($\uparrow$1.8, $\uparrow$5.0)} \\
& & TaylorPODA
& $\uparrow$\textbf{5.4} {\tiny ($\uparrow$3.1, $\uparrow$7.7)} & $\uparrow$\textbf{11.3} {\tiny ($\uparrow$7.1, $\uparrow$16.4)}
& $\uparrow$\textbf{1.0} {\tiny ($\uparrow$0.2, $\uparrow$2.3)} & $\uparrow$\textbf{5.6} {\tiny ($\uparrow$3.4, $\uparrow$7.7)} \\
\cmidrule(lr){2-7}
& \multirow{4}{*}{\rotatebox{90}{Titanic}}
  & OCC-1 
& $\uparrow$20.9 {\tiny ($\uparrow$13.7, $\uparrow$27.0)} & $\uparrow$24.2 {\tiny ($\uparrow$9.5, $\uparrow$40.8)}
& $\uparrow$4.3 {\tiny ($\uparrow$1.0, $\uparrow$8.7)} & $\uparrow$13.6 {\tiny ($\uparrow$8.7, $\uparrow$19.4)} \\
& & LIME
& $\downarrow$12.3 {\tiny ($\downarrow$17.7, $\downarrow$8.0)} & $\downarrow$7.0 {\tiny ($\downarrow$12.6, $\downarrow$1.1)}
& $\downarrow$3.1 {\tiny ($\downarrow$5.0, $\downarrow$1.7)} & $\downarrow$2.7 {\tiny ($\downarrow$4.2, $\downarrow$1.4)} \\
& & WeightedSHAP
& $\uparrow$\textbf{23.5} {\tiny ($\uparrow$16.9, $\uparrow$29.1)} & $\uparrow$\textbf{45.2} {\tiny ($\uparrow$32.8, $\uparrow$57.0)}
& $\uparrow$\textbf{5.0} {\tiny ($\uparrow$2.5, $\uparrow$8.9)} & $\uparrow$\textbf{13.7} {\tiny ($\uparrow$8.8, $\uparrow$19.4)} \\
& & TaylorPODA
& $\uparrow$7.6 {\tiny ($\uparrow$6.0, $\uparrow$9.4)} & $\uparrow$16.5 {\tiny ($\uparrow$14.5, $\uparrow$18.8)}
& $\uparrow$0.6 {\tiny ($\uparrow$0.0, $\uparrow$1.2)} & $\uparrow$2.6 {\tiny ($\uparrow$1.4, $\uparrow$3.8)} \\

\cmidrule(lr){1-7}
\multirow{12}{*}{\rotatebox{90}{XGBoost}}
& \multirow{4}{*}{\rotatebox{90}{Cancer}}
  & OCC-1 
& $\downarrow$66.5 {\tiny ($\downarrow$89.2, $\downarrow$48.7)} & $\downarrow$64.3 {\tiny ($\downarrow$87.7, $\downarrow$42.6)}
& $\downarrow$9.1 {\tiny ($\downarrow$12.3, $\downarrow$6.7)} & $\downarrow$8.6 {\tiny ($\downarrow$12.5, $\downarrow$4.9)} \\
& & LIME
& $\downarrow$12.2 {\tiny ($\downarrow$20.6, $\downarrow$6.7)} & $\downarrow$32.5 {\tiny ($\downarrow$42.0, $\downarrow$25.0)}
& $\downarrow$0.9 {\tiny ($\downarrow$2.1, $\downarrow$0.2)} & $\downarrow$3.6 {\tiny ($\downarrow$5.4, $\downarrow$1.9)} \\
& & WeightedSHAP
& $\uparrow$\textbf{3.6} {\tiny ($\uparrow$1.1, $\uparrow$7.4)} & $\uparrow$\textbf{12.0} {\tiny ($\uparrow$6.3, $\uparrow$21.2)}
& $\uparrow$0.1 {\tiny ($\uparrow$0.0, $\uparrow$0.4)} & $\uparrow$\textbf{1.6} {\tiny ($\uparrow$0.6, $\uparrow$3.3)} \\
& & TaylorPODA
& $\uparrow$3.5 {\tiny ($\uparrow$0.6, $\uparrow$7.6)} & $\uparrow$9.5 {\tiny ($\uparrow$5.3, $\uparrow$15.2)}
& $\uparrow$\textbf{0.4} {\tiny ($\uparrow$0.0, $\uparrow$1.1)} & $\uparrow$\textbf{1.6} {\tiny ($\uparrow$0.8, $\uparrow$2.7)} \\
\cmidrule(lr){2-7}
& \multirow{4}{*}{\rotatebox{90}{Rice}}
  & OCC-1 
& $\downarrow$5.3 {\tiny ($\downarrow$15.1, $\uparrow$4.4)} & $\downarrow$9.2 {\tiny ($\downarrow$22.1, $\uparrow$3.3)}
& $\uparrow$1.7 {\tiny ($\downarrow$0.4, $\uparrow$4.3)} & $\uparrow$1.9 {\tiny ($\downarrow$1.0, $\uparrow$4.7)} \\
& & LIME
& $\downarrow$7.3 {\tiny ($\downarrow$15.3, $\uparrow$0.5)} & $\downarrow$27.2 {\tiny ($\downarrow$37.3, $\downarrow$16.9)}
& $\downarrow$1.3 {\tiny ($\downarrow$3.1, $\uparrow$0.8)} & $\downarrow$3.0 {\tiny ($\downarrow$5.6, $\uparrow$0.4)} \\
& & WeightedSHAP
& $\uparrow$\textbf{9.0} {\tiny ($\uparrow$3.3, $\uparrow$18.6)} & $\uparrow$\textbf{14.0} {\tiny ($\uparrow$6.9, $\uparrow$24.3)}
& $\uparrow$\textbf{2.3} {\tiny ($\uparrow$0.7, $\uparrow$4.6)} & $\uparrow$4.3 {\tiny ($\uparrow$1.8, $\uparrow$6.7)} \\
& & TaylorPODA
& $\uparrow$5.7 {\tiny ($\uparrow$1.5, $\uparrow$12.6)} & $\uparrow$8.3 {\tiny ($\uparrow$3.7, $\uparrow$13.9)}
& $\uparrow$0.9 {\tiny ($\uparrow$0.1, $\uparrow$1.8)} & $\uparrow$\textbf{4.6} {\tiny ($\uparrow$2.2, $\uparrow$6.7)} \\
\cmidrule(lr){2-7}
& \multirow{4}{*}{\rotatebox{90}{Titanic}}
  & OCC-1 
& $\uparrow$55.1 {\tiny ($\uparrow$40.0, $\uparrow$69.0)} & $\uparrow$59.7 {\tiny ($\uparrow$42.8, $\uparrow$74.7)}
& $\uparrow$11.6 {\tiny ($\uparrow$6.5, $\uparrow$16.8)} & $\uparrow$\textbf{15.6} {\tiny ($\uparrow$11.1, $\uparrow$21.1)} \\
& & LIME
& $\downarrow$25.0 {\tiny ($\downarrow$34.1, $\downarrow$17.5)} & $\downarrow$29.6 {\tiny ($\downarrow$37.3, $\downarrow$22.2)}
& $\downarrow$3.7 {\tiny ($\downarrow$6.9, $\downarrow$1.3)} & $\downarrow$4.3 {\tiny ($\downarrow$6.1, $\downarrow$2.3)} \\
& & WeightedSHAP
& $\uparrow$\textbf{56.1} {\tiny ($\uparrow$41.5, $\uparrow$69.9)} & $\uparrow$\textbf{64.9} {\tiny ($\uparrow$50.9, $\uparrow$78.8)}
& $\uparrow$\textbf{11.6} {\tiny ($\uparrow$6.5, $\uparrow$16.8)} & $\uparrow$\textbf{15.6} {\tiny ($\uparrow$11.1, $\uparrow$21.1)} \\
& & TaylorPODA
& $\uparrow$20.1 {\tiny ($\uparrow$14.6, $\uparrow$24.0)} & $\uparrow$21.6 {\tiny ($\uparrow$17.3, $\uparrow$25.3)}
& $\uparrow$4.3 {\tiny ($\uparrow$1.9, $\uparrow$6.3)} & $\uparrow$3.9 {\tiny ($\uparrow$2.4, $\uparrow$5.4)} \\

\bottomrule
\end{tabular}
\end{center}
\label{table_adaptation_cls}
\end{table}
\section{Empirical analysis: toward user-defined utilities}

\subsection{Experimental instantiation of TaylorPODA}

TaylorPODA defines a feasible allocation space for distributing Taylor interaction effects by enforcing
$
\sum_{i\in S}\xi_{i,S}=1
$
by construction. Notably, this formulation admits the Shapley-value allocation as a special feasible solution. Specifically, the uniform allocation $\xi_{i,S}=1/|S|$ for all $i\in S$ satisfies the constraints of TaylorPODA. Consequently, TaylorPODA can be viewed as generalizing the attribution mechanism of the Shapley value by allowing a broader family of interaction-effect allocations.

For empirical evaluation, we further instantiate this feasible allocation space using a Dirichlet-based sampling strategy. The Dirichlet distribution naturally satisfies the constraint $\sum_{i\in S}\xi_{i,S}=1$ while generating diverse feasible allocations. Let $S=\{r_1,\ldots,r_{|S|}\}$. Given a parameter vector $\boldsymbol{\alpha}_S=(\alpha_{r_1,S},\dots,\alpha_{r_{|S|},S})$, a sample $\boldsymbol{\xi}_S=(\xi_{r_1,S},\dots,\xi_{r_{|S|},S})$ is drawn from the Dirichlet distribution with density
\begin{equation} 
\label{eqn_dirichlet} 
    \begin{aligned} 
    p(\boldsymbol{\xi}_S;\boldsymbol{\alpha}_S)
    =
    \frac{
    \Gamma\!\left(\sum_{j=1}^{|S|}\alpha_{r_j,S}\right)
    }{
    \prod_{j=1}^{|S|}\Gamma(\alpha_{r_j,S})
    }
    \prod_{j=1}^{|S|}
    \xi_{r_j,S}^{\alpha_{r_j,S}-1},
    \end{aligned} 
\end{equation} 
where $\Gamma(\cdot)$ is the Gamma function. The parameter vector $\boldsymbol{\alpha}_S$ controls the concentration of the sampled allocations. Further properties of the Dirichlet distribution can be found in \cite{ng2011dirichlet}. Throughout our experiments, we set $\alpha_{i,S}=1$ for all $S\subseteq N$ and $i\in S$, and evaluate 16 feasible allocations per instance (matching the default number of candidate weight sets used in WeightedSHAP). The candidate allocations comprise the Shapley-value allocation and 15 additional allocations sampled from the Dirichlet distribution. 

Additional implementation details for the experiments in this section are provided in Appendix~\ref{apdx_implementation}. 

\subsection{Primary evaluation: quantitative performance under utility objectives}
We empirically evaluate the effectiveness of TaylorPODA across multiple utility functions. For classification tasks, we adopt the Inclusion Area Under the top-1 Accuracy Curve (Inclusion AUC) and Exclusion AUC proposed by \cite{jethani2022fastshap}, which evaluate attribution rankings through progressive feature inclusion and exclusion. For regression tasks, we consider Inclusion MSE~\citep{kwonWeightedSHAPAnalyzingImproving2022}, together with a corresponding Exclusion MSE metric analogously constructed through progressive exclusion of highly attributed features. Following the same evaluation protocol, we further adopt the Area Under the Prediction Recovery Curve (AUP)~\citep{kwonWeightedSHAPAnalyzingImproving2022}, which we refer to as Inclusion AUP, along with a corresponding Exclusion AUP metric analogously defined through progressive feature exclusion. Inclusion AUP and Exclusion AUP are applicable to both classification and regression tasks. Collectively, these utility functions define diverse evaluation settings through which users may assess the “goodness” of the explanation with different downstream preferences. Moreover, we use the Shapley value as the primary baseline, since it satisfies all the proposed postulates, as discussed in Section~\ref{s_theoretical_analysis}. Specifically, we report the mean improvement in utility-function values relative to the Shapley value over 100 hold-out test instances, together with 95\% confidence intervals (CIs) estimated from 100 bootstrap resamples. For classification tasks, we use the Cancer~\citep{cancer}, Rice~\citep{rice}, and Titanic~\citep{titanic} datasets. For regression tasks, we use the Abalone~\citep{abalone}, California~\citep{california}, and Concrete~\citep{concrete} datasets. The evaluated models include fully differentiable multilayer perceptrons (MLPs) with \texttt{tanh} and \texttt{logistic} activations, partially non-differentiable MLPs with \texttt{ReLU} activations, and the non-differentiable tree-based model XGBoost~\citep{chen2016xgboost}.

As shown in Tables~\ref{table_adaptation_cls} and \ref{table_adaptation_rgr}, \textbf{TaylorPODA consistently achieves utility improvements over the Shapley value across all evaluated utility functions for both classification and regression tasks}. These results demonstrate the effectiveness of the proposed \textit{adaptation} mechanism in aligning attribution results with the specified utility objectives while preserving a principled attribution process that satisfies all the proposed postulates. Notably, the observed improvements remain consistent across fully differentiable, partially non-differentiable, and non-differentiable models. These improvements suggest that the underlying attribution rationale extends effectively beyond the differentiability assumptions of the Taylor expansion framework, as discussed in Section~\ref{s_theoretical_analysis}. While WeightedSHAP achieves the best performance most often, TaylorPODA consistently ranks among the strongest-performing methods and in several cases achieves superior results. In contrast, OCC-1 and LIME do not exhibit a consistent performance advantage over the Shapley value across the evaluated objectives and datasets. More importantly, TaylorPODA maintains such competitive performance while preserving all the proposed postulates, thereby retaining a well-grounded mechanism. 
\renewcommand{\arraystretch}{1.2}
\begin{table}[htp]
\caption{Adaptation performance under different user-defined utility objectives on regression tasks. Values denote differences from the Shapley value baseline. ($\uparrow$ and $\downarrow$ indicate changes toward better and worse performance, respectively.)}
\begin{center}
\footnotesize
\begin{tabular}{cc c llll}
\toprule
\makecell{\textbf{Model}} & \makecell{\textbf{Data}} & \makecell{\textbf{Method}} & \makecell{\textbf{Inclusion AUP}\\\tiny{($\times10^{-2}$)}} & \makecell{\textbf{Exclusion AUP}\\\tiny{($\times10^{-2}$)}} & \makecell{\textbf{Inclusion MSE}\\\tiny{($\times10^{-4}$)}} & \makecell{\textbf{Exclusion MSE}\\\tiny{($\times10^{-4}$)}}\\
\midrule

\multirow{12}{*}{\rotatebox{90}{\makecell{MLP \tiny{(\texttt{tanh} \& \texttt{logistic})}}}}
& \multirow{4}{*}{\rotatebox{90}{Abalone}} 
  & OCC-1 
& $\uparrow$1.0 {\tiny ($\downarrow$0.4, $\uparrow$2.5)} & $\downarrow$0.9 {\tiny ($\downarrow$3.3, $\uparrow$1.3)}
& $\downarrow$1.4 {\tiny ($\downarrow$2.9, $\downarrow$0.1)} & $\downarrow$2.9 {\tiny ($\downarrow$6.5, $\uparrow$2.0)} \\
& & LIME
& $\uparrow$2.1 {\tiny ($\uparrow$0.9, $\uparrow$3.5)} & $\downarrow$4.6 {\tiny ($\downarrow$5.7, $\downarrow$3.3)}
& $\uparrow$0.7 {\tiny ($\downarrow$0.4, $\uparrow$2.0)} & $\downarrow$5.2 {\tiny ($\downarrow$6.8, $\downarrow$3.1)} \\
& & WeightedSHAP
& $\uparrow$6.1 {\tiny ($\uparrow$5.0, $\uparrow$7.5)} & $\uparrow$\textbf{7.1} {\tiny ($\uparrow$5.8, $\uparrow$9.1)}
& $\uparrow$3.3 {\tiny ($\uparrow$2.4, $\uparrow$4.5)} & $\uparrow$\textbf{8.0} {\tiny ($\uparrow$4.8, $\uparrow$13.2)} \\
& & TaylorPODA
& $\uparrow$\textbf{6.6} {\tiny ($\uparrow$5.4, $\uparrow$7.5)} & $\uparrow$4.8 {\tiny ($\uparrow$4.0, $\uparrow$6.0)}
& $\uparrow$\textbf{4.1} {\tiny ($\uparrow$3.0, $\uparrow$5.5)} & $\uparrow$5.0 {\tiny ($\uparrow$3.6, $\uparrow$7.4)} \\
\cmidrule(lr){2-7}
& \multirow{4}{*}{\rotatebox{90}{California}} 
  & OCC-1 
& $\downarrow$0.6 {\tiny ($\downarrow$4.5, $\uparrow$2.7)} & $\downarrow$1.2 {\tiny ($\downarrow$4.8, $\uparrow$2.3)}
& $\downarrow$12.0 {\tiny ($\downarrow$25.2, $\downarrow$0.9)} & $\downarrow$7.2 {\tiny ($\downarrow$20.9, $\uparrow$7.9)} \\
& & LIME
& $\downarrow$8.3 {\tiny ($\downarrow$10.7, $\downarrow$5.7)} & $\downarrow$7.0 {\tiny ($\downarrow$9.0, $\downarrow$4.4)}
& $\downarrow$18.3 {\tiny ($\downarrow$25.1, $\downarrow$11.0)} & $\downarrow$16.1 {\tiny ($\downarrow$21.1, $\downarrow$9.7)} \\
& & WeightedSHAP
& $\uparrow$\textbf{5.9} {\tiny ($\uparrow$4.4, $\uparrow$7.9)} & $\uparrow$\textbf{7.5} {\tiny ($\uparrow$6.0, $\uparrow$9.7)}
& $\uparrow$5.4 {\tiny ($\uparrow$3.1, $\uparrow$8.6)} & $\uparrow$\textbf{28.1} {\tiny ($\uparrow$20.2, $\uparrow$39.6)} \\
& & TaylorPODA
& $\uparrow$3.5 {\tiny ($\uparrow$2.7, $\uparrow$4.6)} & $\uparrow$6.8 {\tiny ($\uparrow$5.6, $\uparrow$7.7)}
& $\uparrow$\textbf{3.2} {\tiny ($\uparrow$2.1, $\uparrow$5.0)} & $\uparrow$22.7 {\tiny ($\uparrow$16.9, $\uparrow$29.5)} \\
\cmidrule(lr){2-7}
& \multirow{4}{*}{\rotatebox{90}{Concrete}} 
  & OCC-1 
& $\downarrow$3.8 {\tiny ($\downarrow$6.4, $\downarrow$1.8)} & $\downarrow$2.5 {\tiny ($\downarrow$6.4, $\uparrow$2.0)}
& $\downarrow$9.6 {\tiny ($\downarrow$17.4, $\downarrow$3.4)} & $\downarrow$4.7 {\tiny ($\downarrow$25.5, $\uparrow$10.7)} \\
& & LIME
& $\downarrow$10.1 {\tiny ($\downarrow$12.5, $\downarrow$8.2)} & $\downarrow$6.7 {\tiny ($\downarrow$8.9, $\downarrow$3.7)}
& $\downarrow$17.5 {\tiny ($\downarrow$22.3, $\downarrow$11.8)} & $\downarrow$15.5 {\tiny ($\downarrow$26.0, $\uparrow$1.6)} \\
& & WeightedSHAP
& $\uparrow$\textbf{4.9} {\tiny ($\uparrow$3.6, $\uparrow$6.5)} & $\uparrow$\textbf{9.8} {\tiny ($\uparrow$7.6, $\uparrow$12.2)}
& $\uparrow$6.0 {\tiny ($\uparrow$3.8, $\uparrow$8.8)} & $\uparrow$\textbf{33.3} {\tiny ($\uparrow$25.6, $\uparrow$42.5)} \\
& & TaylorPODA
& $\uparrow$4.8 {\tiny ($\uparrow$4.0, $\uparrow$5.7)} & $\uparrow$9.2 {\tiny ($\uparrow$7.9, $\uparrow$10.3)}
& $\uparrow$\textbf{6.5} {\tiny ($\uparrow$5.0, $\uparrow$8.4)} & $\uparrow$33.1 {\tiny ($\uparrow$25.5, $\uparrow$40.0)} \\

\cmidrule(lr){1-7}
\multirow{12}{*}{\rotatebox{90}{\makecell{MLP \tiny{(\texttt{ReLU})}}}}
& \multirow{4}{*}{\rotatebox{90}{Abalone}} 
  & OCC-1 
& $\downarrow$0.3 {\tiny ($\downarrow$1.6, $\uparrow$0.9)} & $\downarrow$0.7 {\tiny ($\downarrow$3.1, $\uparrow$1.1)}
& $\downarrow$2.0 {\tiny ($\downarrow$3.2, $\downarrow$0.9)} & $\downarrow$3.3 {\tiny ($\downarrow$6.3, $\downarrow$0.6)} \\
& & LIME
& $\uparrow$1.1 {\tiny ($\uparrow$0.1, $\uparrow$2.5)} & $\downarrow$4.6 {\tiny ($\downarrow$6.2, $\downarrow$2.9)}
& $\downarrow$0.2 {\tiny ($\downarrow$1.4, $\uparrow$0.7)} & $\downarrow$5.6 {\tiny ($\downarrow$8.5, $\downarrow$2.9)} \\
& & WeightedSHAP
& $\uparrow$5.3 {\tiny ($\uparrow$4.3, $\uparrow$6.4)} & $\uparrow$\textbf{7.1} {\tiny ($\uparrow$6.1, $\uparrow$8.7)}
& $\uparrow$2.9 {\tiny ($\uparrow$2.0, $\uparrow$3.9)} & $\uparrow$\textbf{7.9} {\tiny ($\uparrow$5.3, $\uparrow$12.6)} \\
& & TaylorPODA
& $\uparrow$\textbf{6.6} {\tiny ($\uparrow$5.5, $\uparrow$7.7)} & $\uparrow$4.4 {\tiny ($\uparrow$3.5, $\uparrow$5.1)}
& $\uparrow$\textbf{3.8} {\tiny ($\uparrow$2.8, $\uparrow$4.9)} & $\uparrow$5.6 {\tiny ($\uparrow$3.7, $\uparrow$9.0)} \\
\cmidrule(lr){2-7}
& \multirow{4}{*}{\rotatebox{90}{California}} 
  & OCC-1 
& $\downarrow$1.6 {\tiny ($\downarrow$5.3, $\uparrow$1.8)} & $\downarrow$0.7 {\tiny ($\downarrow$4.6, $\uparrow$3.1)}
& $\downarrow$13.6 {\tiny ($\downarrow$27.0, $\downarrow$2.9)} & $\downarrow$3.4 {\tiny ($\downarrow$18.8, $\uparrow$12.5)} \\
& & LIME
& $\downarrow$8.4 {\tiny ($\downarrow$10.5, $\downarrow$5.7)} & $\downarrow$6.6 {\tiny ($\downarrow$8.6, $\downarrow$4.2)}
& $\downarrow$18.4 {\tiny ($\downarrow$25.2, $\downarrow$11.6)} & $\downarrow$13.7 {\tiny ($\downarrow$20.1, $\downarrow$7.3)} \\
& & WeightedSHAP
& $\uparrow$\textbf{5.5} {\tiny ($\uparrow$3.8, $\uparrow$7.7)} & $\uparrow$\textbf{7.6} {\tiny ($\uparrow$5.9, $\uparrow$10.1)}
& $\uparrow$\textbf{5.6} {\tiny ($\uparrow$3.2, $\uparrow$9.0)} & $\uparrow$\textbf{29.0} {\tiny ($\uparrow$19.8, $\uparrow$41.1)} \\
& & TaylorPODA
& $\uparrow$3.6 {\tiny ($\uparrow$2.8, $\uparrow$4.9)} & $\uparrow$6.0 {\tiny ($\uparrow$5.1, $\uparrow$6.8)}
& $\uparrow$3.8 {\tiny ($\uparrow$2.3, $\uparrow$6.5)} & $\uparrow$19.6 {\tiny ($\uparrow$14.8, $\uparrow$23.8)} \\
\cmidrule(lr){2-7}
& \multirow{4}{*}{\rotatebox{90}{Concrete}} 
  & OCC-1 
& $\downarrow$4.3 {\tiny ($\downarrow$6.6, $\downarrow$1.8)} & $\downarrow$0.2 {\tiny ($\downarrow$3.2, $\uparrow$3.4)}
& $\downarrow$11.5 {\tiny ($\downarrow$19.4, $\downarrow$4.2)} & $\uparrow$5.7 {\tiny ($\downarrow$7.2, $\uparrow$18.2)} \\
& & LIME
& $\downarrow$10.9 {\tiny ($\downarrow$12.9, $\downarrow$8.5)} & $\downarrow$7.1 {\tiny ($\downarrow$9.6, $\downarrow$3.4)}
& $\downarrow$18.6 {\tiny ($\downarrow$22.4, $\downarrow$13.9)} & $\downarrow$21.6 {\tiny ($\downarrow$32.4, $\downarrow$4.3)} \\
& & WeightedSHAP
& $\uparrow$\textbf{4.9} {\tiny ($\uparrow$3.6, $\uparrow$6.6)} & $\uparrow$\textbf{10.1} {\tiny ($\uparrow$8.4, $\uparrow$12.1)}
& $\uparrow$6.0 {\tiny ($\uparrow$3.7, $\uparrow$8.6)} & $\uparrow$\textbf{34.0} {\tiny ($\uparrow$26.6, $\uparrow$45.3)} \\
& & TaylorPODA
& $\uparrow$4.5 {\tiny ($\uparrow$3.5, $\uparrow$5.2)} & $\uparrow$8.9 {\tiny ($\uparrow$7.8, $\uparrow$10.3)}
& $\uparrow$\textbf{6.3} {\tiny ($\uparrow$4.7, $\uparrow$8.4)} & $\uparrow$28.7 {\tiny ($\uparrow$23.0, $\uparrow$35.0)} \\

\cmidrule(lr){1-7}
\multirow{12}{*}{\rotatebox{90}{XGBoost}}
& \multirow{4}{*}{\rotatebox{90}{Abalone}} 
  & OCC-1 
& $\uparrow$1.1 {\tiny ($\downarrow$1.0, $\uparrow$2.6)} & $\uparrow$1.4 {\tiny ($\downarrow$0.8, $\uparrow$3.3)}
& $\downarrow$0.5 {\tiny ($\downarrow$2.7, $\uparrow$1.1)} & $\downarrow$0.4 {\tiny ($\downarrow$4.0, $\uparrow$3.0)} \\
& & LIME
& $\downarrow$3.4 {\tiny ($\downarrow$4.7, $\downarrow$1.8)} & $\downarrow$4.7 {\tiny ($\downarrow$6.3, $\downarrow$2.8)}
& $\downarrow$3.4 {\tiny ($\downarrow$5.1, $\downarrow$1.4)} & $\downarrow$5.7 {\tiny ($\downarrow$7.8, $\downarrow$3.6)} \\
& & WeightedSHAP
& $\uparrow$\textbf{6.1} {\tiny ($\uparrow$5.1, $\uparrow$7.2)} & $\uparrow$\textbf{8.9} {\tiny ($\uparrow$7.4, $\uparrow$10.2)}
& $\uparrow$\textbf{3.8} {\tiny ($\uparrow$3.0, $\uparrow$4.7)} & $\uparrow$\textbf{9.8} {\tiny ($\uparrow$7.4, $\uparrow$12.5)} \\
& & TaylorPODA
& $\uparrow$5.5 {\tiny ($\uparrow$4.4, $\uparrow$7.0)} & $\uparrow$6.5 {\tiny ($\uparrow$4.9, $\uparrow$8.7)}
& $\uparrow$\textbf{3.8} {\tiny ($\uparrow$2.7, $\uparrow$5.4)} & $\uparrow$5.6 {\tiny ($\uparrow$4.4, $\uparrow$7.2)} \\
\cmidrule(lr){2-7}
& \multirow{4}{*}{\rotatebox{90}{California}} 
  & OCC-1 
& $\uparrow$4.6 {\tiny ($\downarrow$0.9, $\uparrow$9.7)} & $\uparrow$7.0 {\tiny ($\uparrow$1.4, $\uparrow$14.3)}
& $\downarrow$0.7 {\tiny ($\downarrow$15.8, $\uparrow$15.6)} & $\uparrow$10.7 {\tiny ($\downarrow$9.3, $\uparrow$39.9)} \\
& & LIME
& $\downarrow$5.8 {\tiny ($\downarrow$8.2, $\downarrow$3.3)} & $\downarrow$1.2 {\tiny ($\downarrow$3.4, $\uparrow$1.0)}
& $\downarrow$13.7 {\tiny ($\downarrow$20.2, $\downarrow$7.9)} & $\downarrow$2.8 {\tiny ($\downarrow$9.0, $\uparrow$3.7)} \\
& & WeightedSHAP
& $\uparrow$\textbf{11.3} {\tiny ($\uparrow$8.4, $\uparrow$15.4)} & $\uparrow$\textbf{14.1} {\tiny ($\uparrow$10.1, $\uparrow$19.9)}
& $\uparrow$\textbf{15.9} {\tiny ($\uparrow$8.8, $\uparrow$28.2)} & $\uparrow$\textbf{41.7} {\tiny ($\uparrow$26.2, $\uparrow$65.8)} \\
& & TaylorPODA
& $\uparrow$4.6 {\tiny ($\uparrow$3.2, $\uparrow$5.9)} & $\uparrow$7.1 {\tiny ($\uparrow$5.8, $\uparrow$8.6)}
& $\uparrow$6.5 {\tiny ($\uparrow$4.3, $\uparrow$9.1)} & $\uparrow$26.3 {\tiny ($\uparrow$19.2, $\uparrow$31.8)} \\
\cmidrule(lr){2-7}
& \multirow{4}{*}{\rotatebox{90}{Concrete}} 
  & OCC-1 
& $\uparrow$10.4 {\tiny ($\uparrow$5.8, $\uparrow$18.2)} & $\uparrow$30.1 {\tiny ($\uparrow$20.0, $\uparrow$46.7)}
& $\uparrow$30.0 {\tiny ($\uparrow$11.7, $\uparrow$66.5)} & $\uparrow$152.8 {\tiny ($\uparrow$94.9, $\uparrow$254.0)} \\
& & LIME
& $\downarrow$15.7 {\tiny ($\downarrow$18.4, $\downarrow$13.4)} & $\downarrow$12.4 {\tiny ($\downarrow$15.8, $\downarrow$9.1)}
& $\downarrow$48.8 {\tiny ($\downarrow$60.9, $\downarrow$40.0)} & $\downarrow$61.3 {\tiny ($\downarrow$75.9, $\downarrow$47.0)} \\
& & WeightedSHAP
& $\uparrow$\textbf{16.8} {\tiny ($\uparrow$13.1, $\uparrow$22.9)} & $\uparrow$\textbf{33.2} {\tiny ($\uparrow$23.5, $\uparrow$48.5)}
& $\uparrow$\textbf{45.5} {\tiny ($\uparrow$28.4, $\uparrow$81.0)} & $\uparrow$\textbf{163.3} {\tiny ($\uparrow$105.4, $\uparrow$261.0)} \\
& & TaylorPODA
& $\uparrow$8.6 {\tiny ($\uparrow$6.8, $\uparrow$10.3)} & $\uparrow$10.0 {\tiny ($\uparrow$8.2, $\uparrow$11.4)}
& $\uparrow$17.1 {\tiny ($\uparrow$11.9, $\uparrow$23.4)} & $\uparrow$42.5 {\tiny ($\uparrow$34.5, $\uparrow$50.7)} \\

\bottomrule
\end{tabular}
\end{center}
\label{table_adaptation_rgr}
\end{table}

\subsection{Additional qualitative performance and examples}
\label{ss_additional_evaluation}

Beyond the primary quantitative evaluation of utility performance, we further investigate additional empirical behaviors of TaylorPODA related to explanation consistency, visualization, and communicability. For illustration, Inclusion AUP is used as a representative utility setting throughout this subsection.

\begin{figure}[tb]
    \centering

    \begin{subfigure}{0.32\linewidth}
        \centering
        \includegraphics[width=\linewidth]{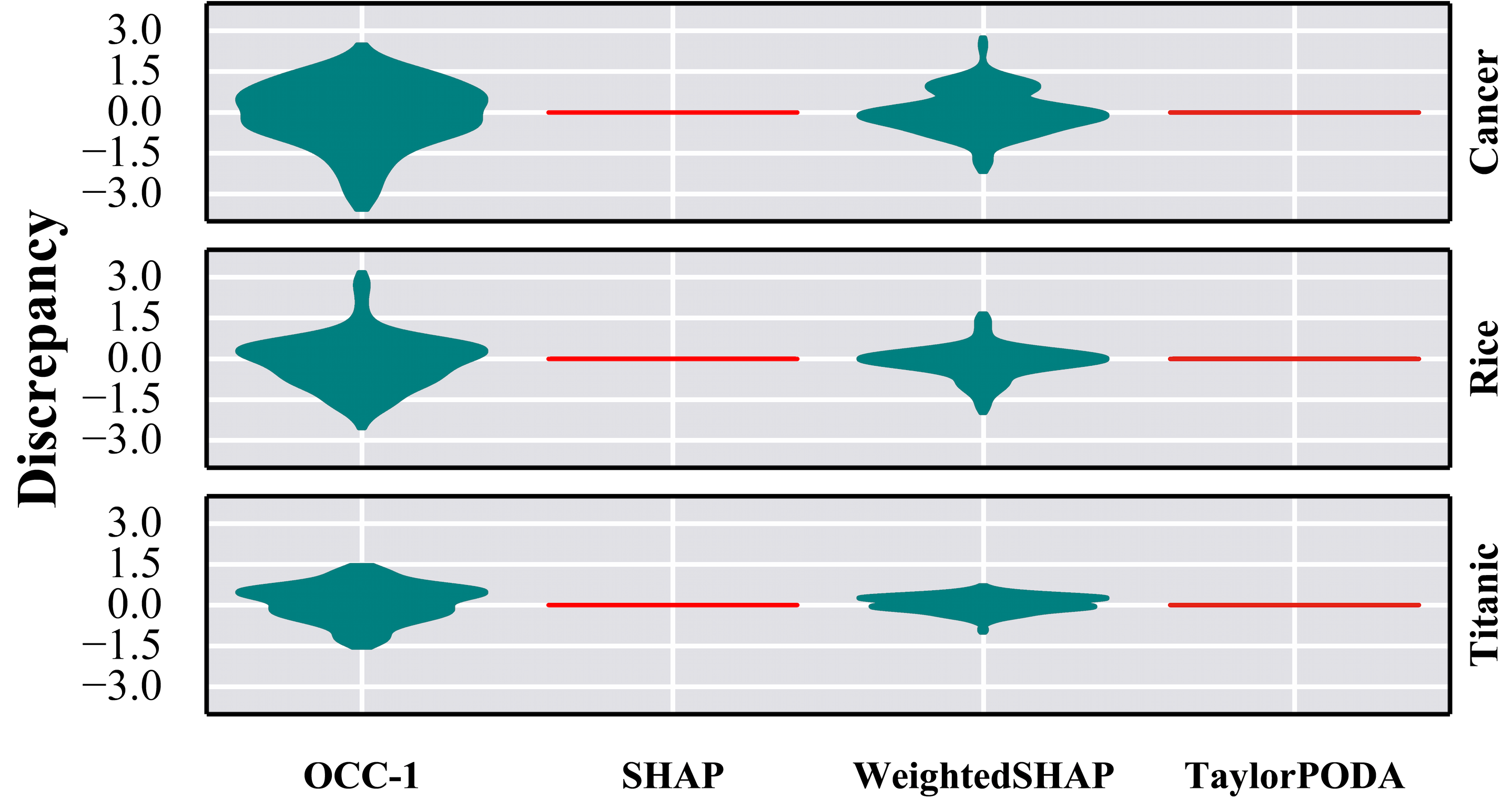}
        \subcaption{\scriptsize{Classification, MLP (\texttt{tanh} \& \texttt{logistic})}}
        \label{fig:d_cls_tanh}
    \end{subfigure}
    \begin{subfigure}{0.32\linewidth}
        \centering
        \includegraphics[width=\linewidth]{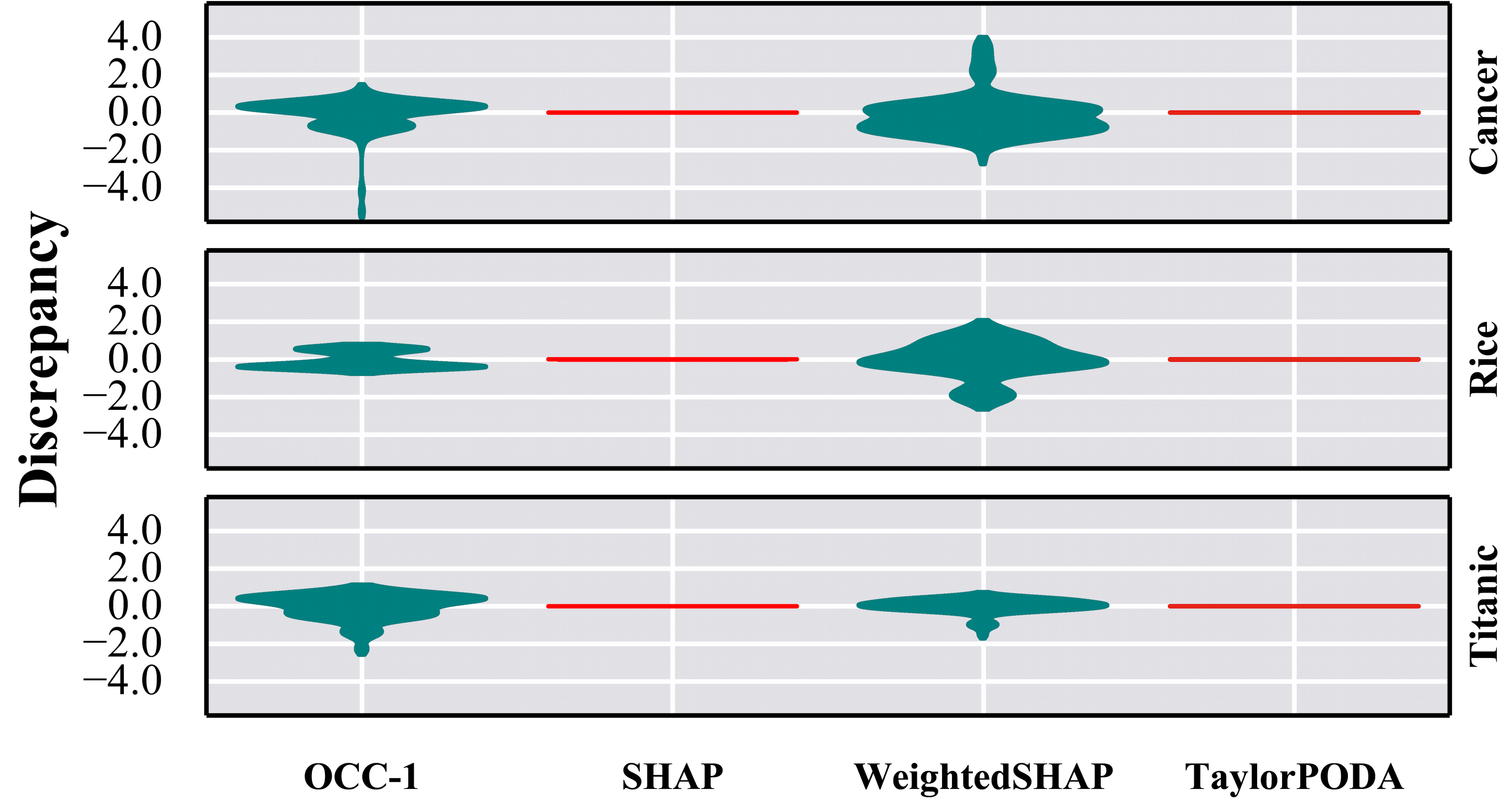}
        \subcaption{\scriptsize{Classification, MLP (\texttt{ReLU})}}
        \label{fig:d_cls_relu}
    \end{subfigure}
    \begin{subfigure}{0.32\linewidth}
        \centering
        \includegraphics[width=\linewidth]{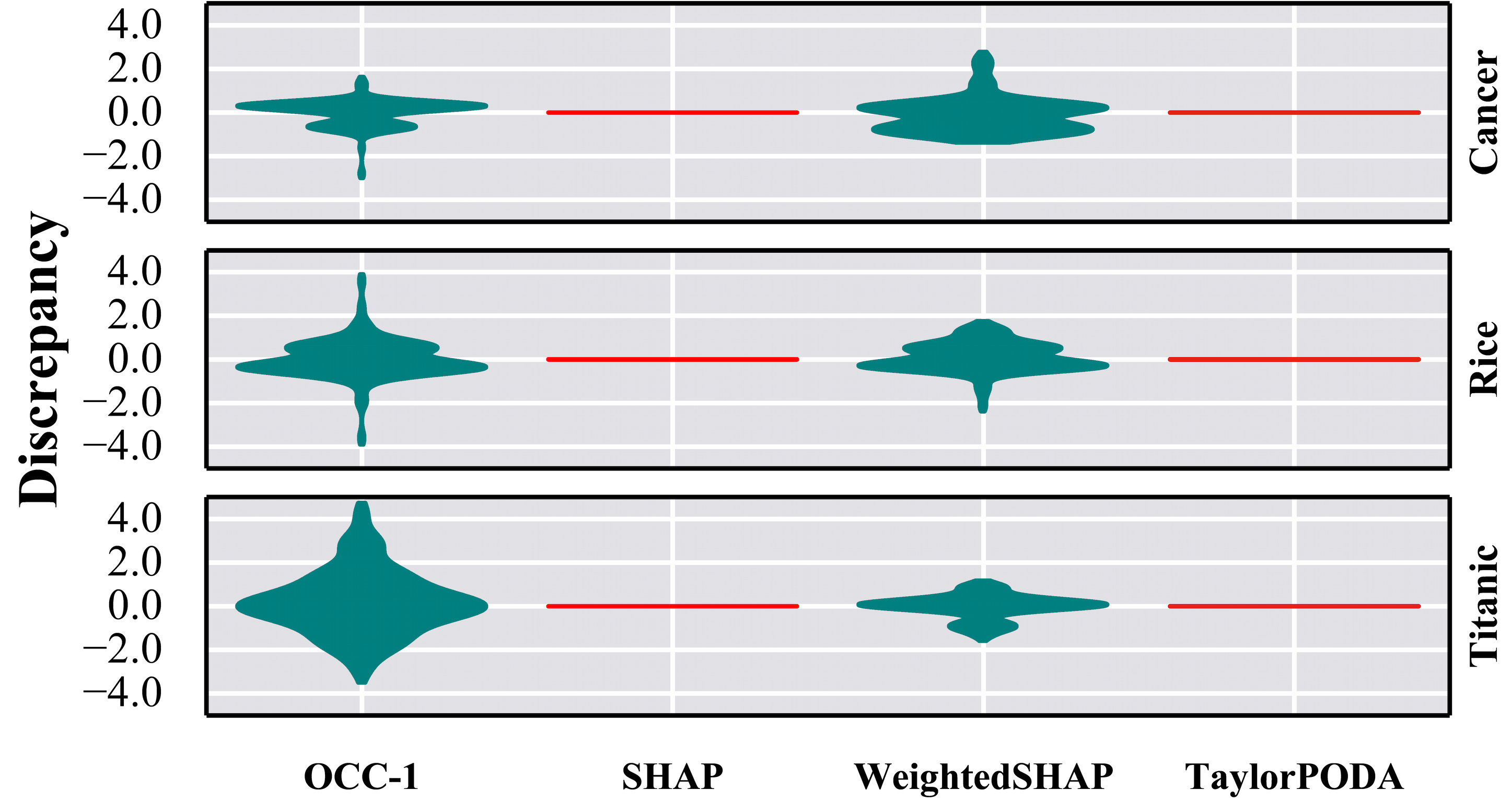}
        \subcaption{\scriptsize{Classification, XGBoost}}
        \label{fig:d_cls_xgb}
    \end{subfigure}
    
    \vspace{1mm}

    \begin{subfigure}{0.32\linewidth}
        \centering
        \includegraphics[width=\linewidth]{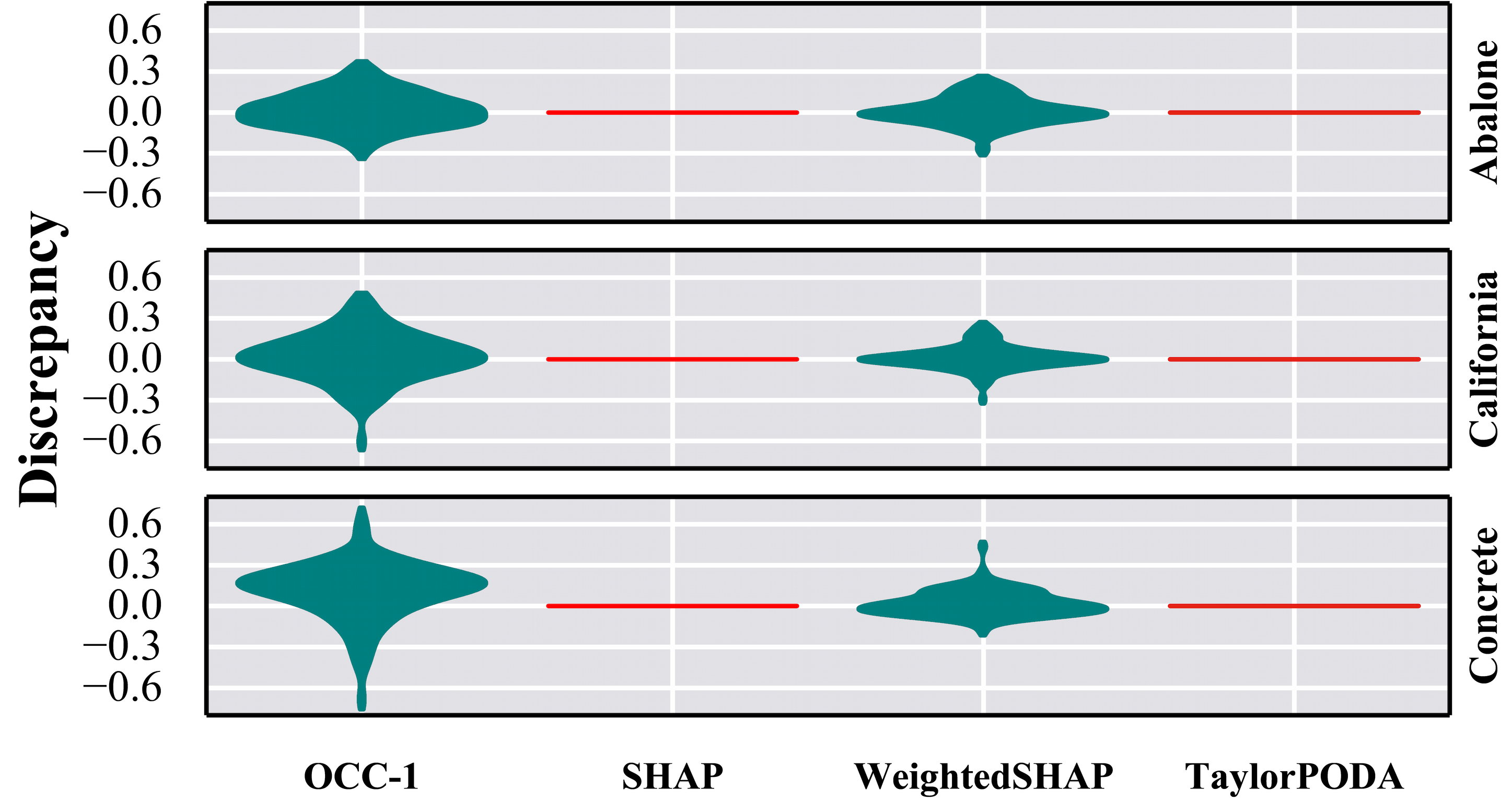}
        \subcaption{\scriptsize{Regression, MLP (\texttt{tanh} \& \texttt{logistic})}}
        \label{fig:d_rgr_tanh}
    \end{subfigure}
    \begin{subfigure}{0.32\linewidth}
        \centering
        \includegraphics[width=\linewidth]{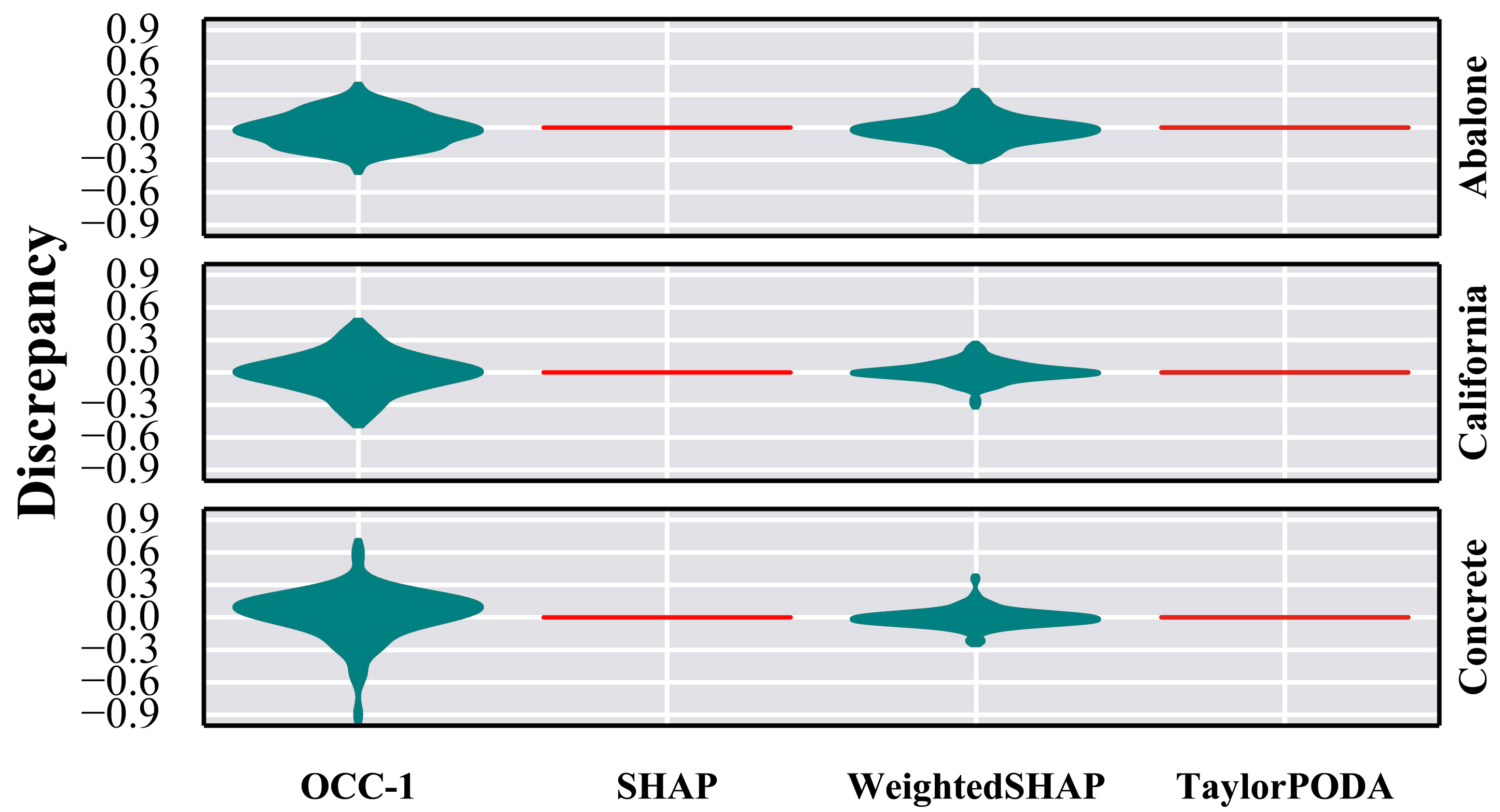}
        \subcaption{\scriptsize{Regression, MLP (\texttt{ReLU})}}
        \label{fig:d_rgr_relu}
    \end{subfigure}
    \begin{subfigure}{0.32\linewidth}
        \centering
        \includegraphics[width=\linewidth]{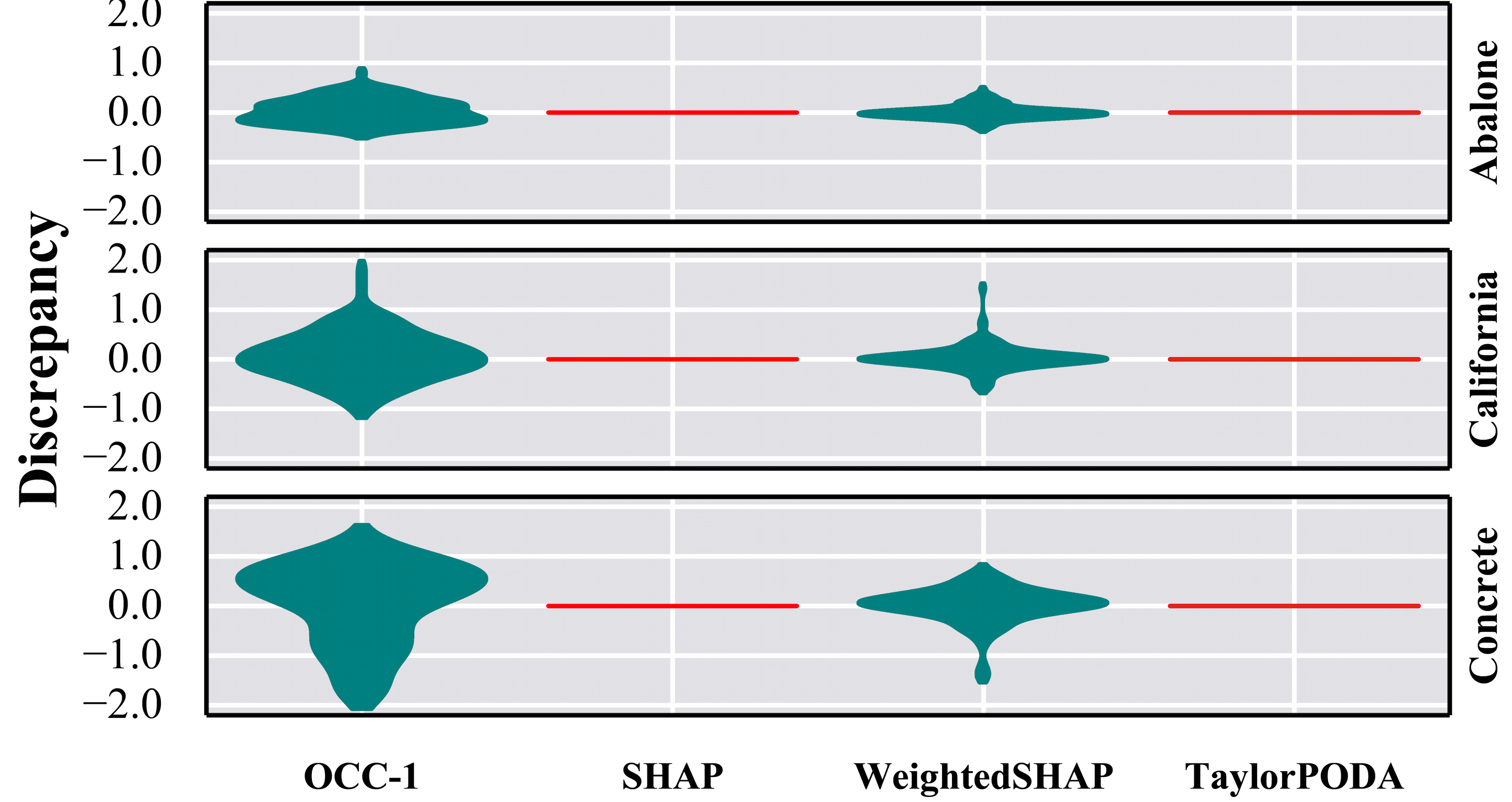}
        \subcaption{\scriptsize{Regression, XGBoost}}
        \label{fig:d_rgr_xgb}
    \end{subfigure}
    \caption{Violin plots of discrepancy values across 100 hold-out test samples per dataset under various models. Both SHAP and TaylorPODA consistently exhibit \textit{zero discrepancy}.}
    \label{figure_discrepancy}
\end{figure}

In particular, we investigate the discrepancy between the model output and the aggregation of feature-wise attribution scores. As shown in the violin plots in Figure~\ref{figure_discrepancy}, \textbf{both TaylorPODA and SHAP consistently exhibit \textit{zero discrepancy} across all test samples}, consistent with the theoretical analysis in Section~\ref{s_theoretical_analysis}. These results suggest that both methods exhaustively allocate Taylor terms (Harsanyi dividends) without redundancy or omission. In contrast, the remaining methods exhibit non-zero discrepancies, where positive values imply overlapping attribution and negative values imply under-allocation. Notably, TaylorPODA maintains \textit{zero discrepancy} even on non-differentiable models, further supporting that its attribution mechanism remains grounded in the exhaustive allocation of Harsanyi dividends beyond the differentiability-based Taylor expansion framework.

\begin{figure}[ht]
\begin{center}
    \begin{minipage}[b]{0.9\linewidth}
        \centering
        \includegraphics[width=\linewidth]{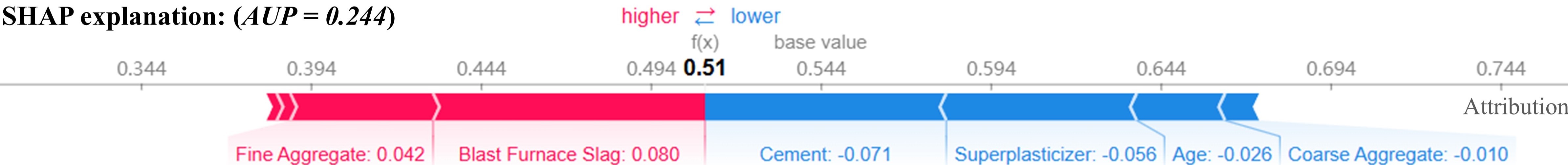}
    \end{minipage}
    
    \begin{minipage}[b]{0.9\linewidth}
        \centering
        \includegraphics[width=\linewidth]{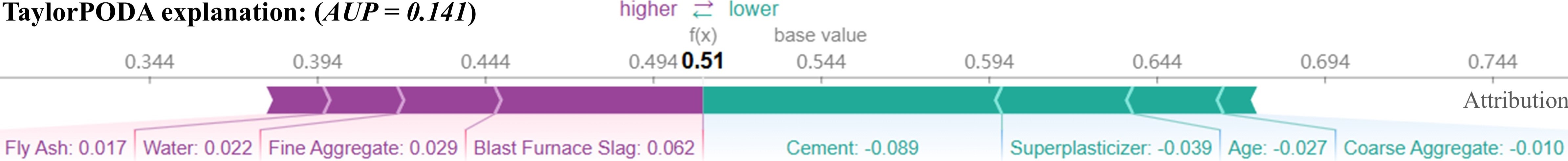}
    \end{minipage}
    \caption{SHAP vs.~TaylorPODA explanations for the same sample from the \textit{Concrete} dataset~\citep{concrete}. Both satisfy the \textit{zero-discrepancy} property, enabling additive bar-plot visualization aligned with the model output. Meanwhile, TaylorPODA achieves a lower (better) \textit{Inclusion AUP}.}    
\label{figure_force_plot}
\end{center}
\end{figure}

Moreover, as illustrated in Figure~\ref{figure_force_plot}, \textbf{TaylorPODA supports SHAP-style visualization by providing feature-wise contributions that remain consistently aligned with the corresponding model outputs for individual predictions}. This enables additive visual explanations in which the aggregation of feature-wise attribution scores together with the base value reconstructs the model output, thereby improving the communicability of the resulting explanation. 

\begin{figure}[tb]
\begin{center}
    \begin{minipage}[b]{0.48\linewidth}
        \centering
        \includegraphics[width=\linewidth]{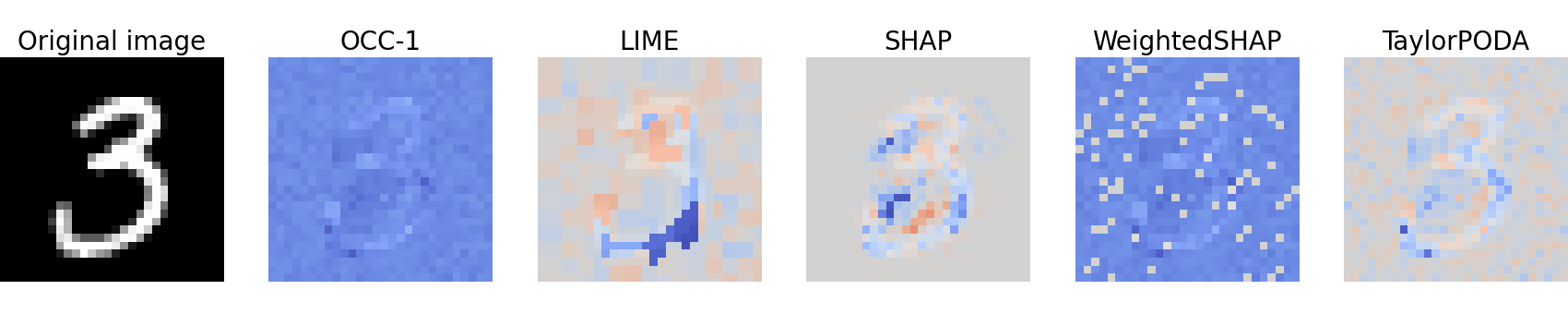}
    \end{minipage}
    \hfill
    \begin{minipage}[b]{0.48\linewidth}
        \centering
        \includegraphics[width=\linewidth]{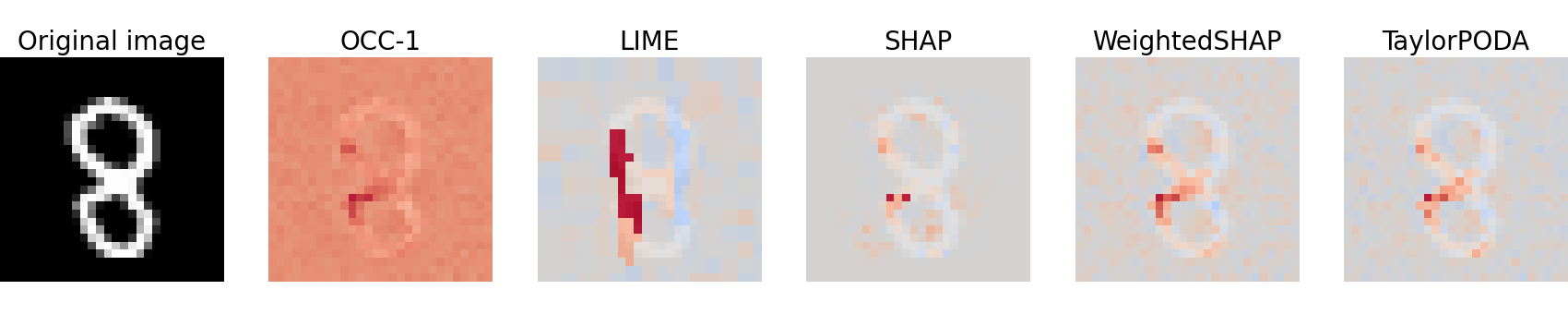}
    \end{minipage}

    \begin{minipage}[b]{0.48\linewidth}
        \centering
        \includegraphics[width=\linewidth]{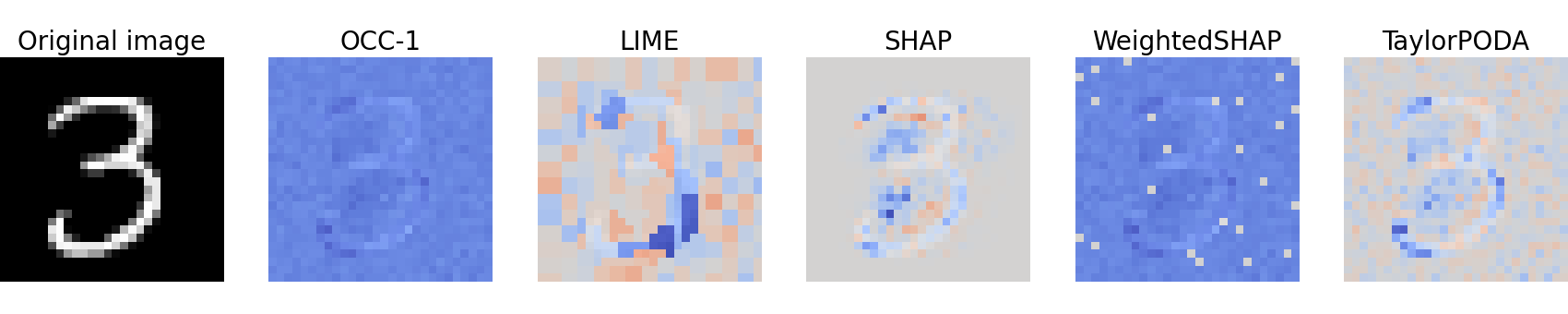}
    \end{minipage}
    \hfill
    \begin{minipage}[b]{0.48\linewidth}
        \centering
        \includegraphics[width=\linewidth]{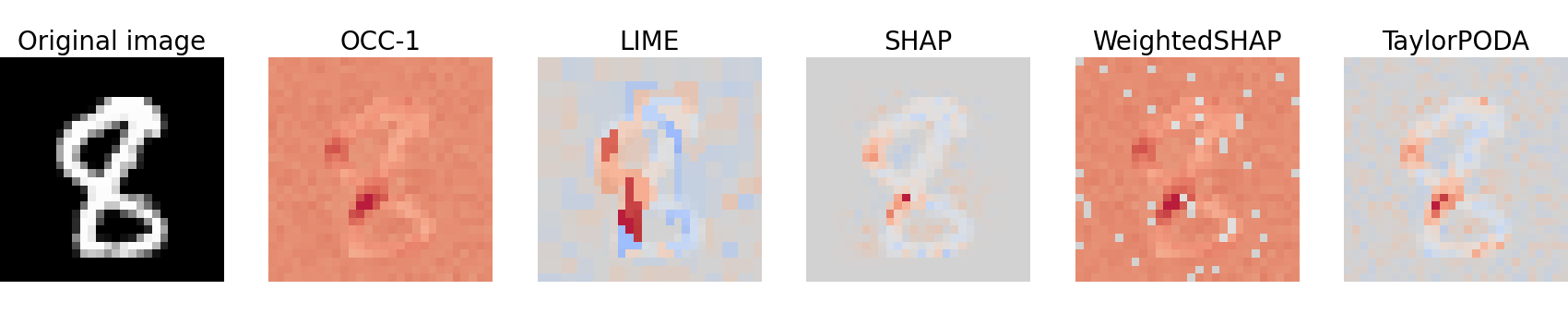}
    \end{minipage}

    \begin{minipage}[b]{0.48\linewidth}
        \centering
        \includegraphics[width=\linewidth]{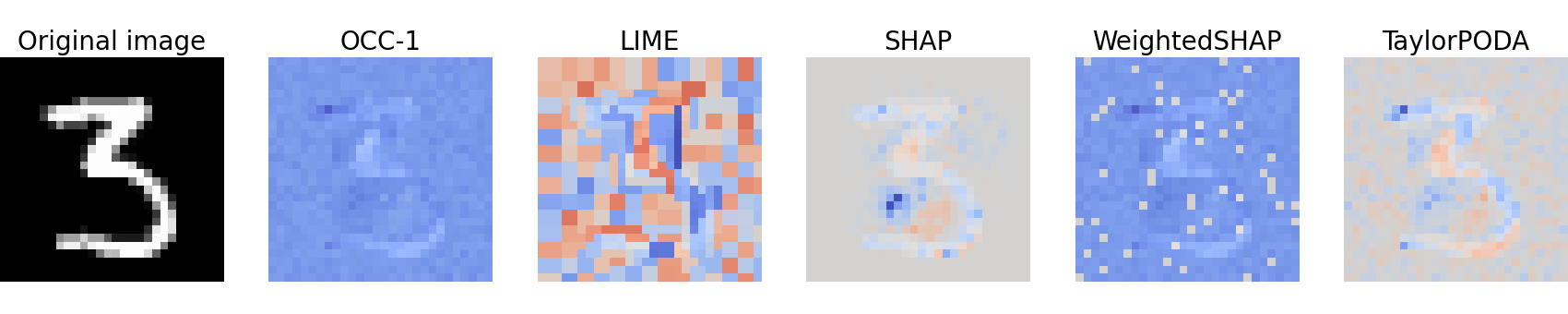}
    \end{minipage}
    \hfill
    \begin{minipage}[b]{0.48\linewidth}
        \centering
        \includegraphics[width=\linewidth]{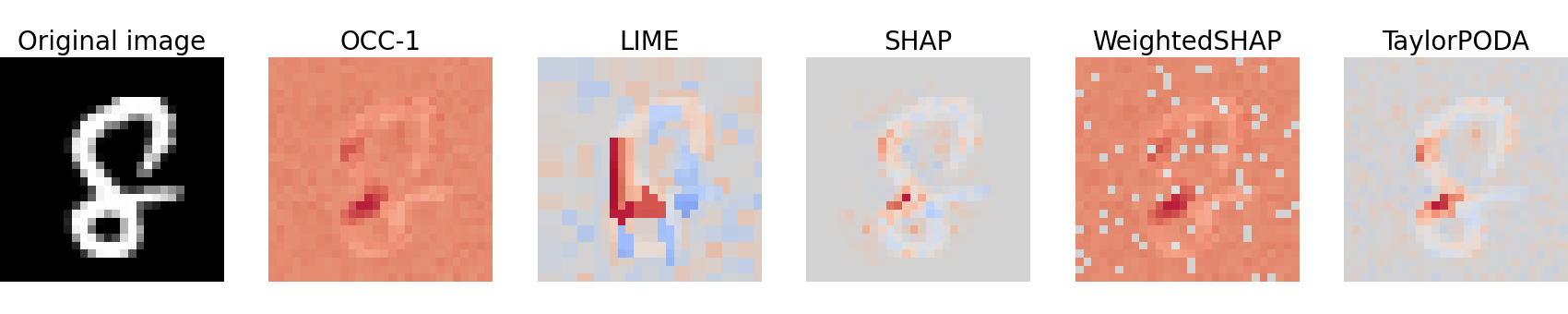}
    \end{minipage}
    \caption{Illustrative examples of attribution explanations for distinguishing digits 3 and 8 in \textit{MNIST}. Pixels are color-coded: blue indicates a negative contribution, whereas red indicates a positive contribution toward predicting “8”.}
    \label{figure_mnist}
\end{center}
\end{figure}

Furthermore, \textbf{TaylorPODA}\footnote{Here, we adopt a heuristic approximation of the fully enumerated version of TaylorPODA in Equation~(\ref{eqn_taylorpoda}) by introducing an upper bound on $|S|$, rather than exhaustively traversing all $S \subseteq N$. Details of this approximation are provided in Appendix~\ref{apdx_approximation}.} \textbf{also produces visually interpretable explanations on image data.} As illustrated in Figure~\ref{figure_mnist} using the \textit{MNIST}~\citep{lecun1998mnist} dataset, the attribution patterns generated by TaylorPODA consistently highlight the openness along the left segments of the upper and lower loops of the digit “8” as key discriminative regions distinguishing it from the digit “3”. Notably, TaylorPODA and SHAP tend to produce visually concentrated attribution patterns with relatively stable attribution ranges. In contrast, methods such as OCC-1 and WeightedSHAP can generate comparatively diffuse or “blurred” attribution patterns, which may require additional filtering or post-processing to better align with human visual perception.

\section{Conclusion and future work}
\label{s_conclusion}

In this paper, we propose TaylorPODA, a new post-hoc model-agnostic method for LA that quantifies feature-wise contributions. TaylorPODA maintains a principled attribution mechanism grounded in postulates derived from the Taylor expansion framework, whereas many existing post-hoc model-agnostic approaches rely on heuristic or only partially justified designs. Moreover, TaylorPODA allows the integration of user-defined utilities into the attribution process, enabling downstream implementations to align explanations with task-specific needs. Within this framework, independent Taylor terms are attributed directly to their corresponding features, while Taylor interaction terms are exhaustively distributed among the involved features with controllable degrees of freedom. Collectively, TaylorPODA provides a more accountable foundation for feature attribution and supports context-adaptive explanations that are responsive to user-specific interpretive needs. This helps avoid the crucial yet frequently overlooked paradox of \textit{explaining opacity with opacity}.

Nonetheless, several important challenges remain open. First, the computational efficiency of TaylorPODA requires further improvement. As defined in Equation~(\ref{eqn_taylorpoda}), a full evaluation requires computing the Harsanyi dividends for all possible feature subsets, resulting in an exponentially large number of masked output queries. Consequently, exhaustive evaluation is currently feasible only for datasets with a relatively small number of input features. For high-dimensional inputs, such as images, the framework relies on approximation schemes to remain computationally tractable. Although the approximations presented in this work already achieve results comparable to SHAP, further methodological advances are needed to ensure well-grounded scalability and efficiency. Second, more refined optimization strategies for the adaptable attribution coefficients remain to be developed. The current Dirichlet-based sampling strategy represents only an initial instantiation. More structured optimization approaches with reduced randomness may further improve adaptation performance and stability across downstream objectives. Third, the design of downstream user-defined utilities itself remains an important research direction. In this work, explanation-fidelity metrics are adopted as representative user-defined utilities for empirical evaluation. Future research may develop utility functions that are more tightly coupled with specific downstream tasks and decision-making scenarios. 

\section*{Acknowledgments}
This work was supported by the UK Research and Innovation (UKRI) Engineering and Physical Sciences Research Council (EPSRC) Doctoral Training Partnership (DTP) through the Healthy Lifespan Institute (HELSI) Flagship Scholarship at the University of Sheffield (Grant No. EP/W524360/1).

\bibliography{ref}
\bibliographystyle{tmlr}

\appendix
\section{Proof of postulate and property satisfaction: TaylorPODA from a Harsanyi-dividend perspective}
\label{apdx_harsanyi_analog}

\textbf{Proof:}

According to the game-theoretic analysis of \cite{grabisch1999axiomatic} and its extension to model outputs in Equation (7) of Appendix A of \cite{deng2024unifying}, we have
\begin{equation}
\label{eqn_fs_apdx}
\begin{aligned}
    f_S(\mathbf{x}) = f(\mathbf{b}) + \sum_{\substack{T\subseteq S\\T\neq\varnothing}}H(T).
\end{aligned}
\end{equation}
Substituting Equation (\ref{eqn_fs_apdx}) into Equation (\ref{eqn_taylorpoda}), we have
\begin{equation}
\label{eqn_taylorpoda_har_apdx}
\begin{aligned}
    \underset{(\text{TaylorPODA})}{a_i}
    &=f(\mathbf{x})-f_{N\backslash\{i\}}(\mathbf{x})-\sum_{\substack{S\subseteq N\\|S|>1\\S\supseteq\{i\}}}(1-\xi_{i, S})H(S)\\
    &=(f(\mathbf{b}) + \sum_{\substack{T\subseteq N\\T\neq\varnothing}}H(T)) - ((f(\mathbf{b}) + \sum_{\substack{T'\subseteq N\backslash\{i\}\\T'\neq\varnothing}}H(T'))) - \sum_{\substack{S\subseteq N\\|S|>1\\S\supseteq\{i\}}}(1-\xi_{i, S})H(S)\\
    &=\sum_{\substack{T\subseteq N\\T\neq\varnothing}}H(T) - \sum_{\substack{T'\subseteq N\backslash\{i\}\\T'\neq\varnothing}}H(T') - \sum_{\substack{S\subseteq N\\|S|>1\\S\supseteq\{i\}}}H(S) + \sum_{\substack{S\subseteq N\\|S|>1\\S\supseteq\{i\}}}\xi_{i, S}H(S)\\
    &=H(\{i\}) + \sum_{\substack{S\subseteq N\\|S|>1\\S\supseteq\{i\}}}\xi_{i, S}H(S).\\
\end{aligned}
\end{equation}

The essence of the Harsanyi dividend is that it captures the additional joint effect arising purely from the act of cooperation itself; for a feature set (coalition) $S$, $H(S)$ represents the intrinsic synergistic contribution generated only when the features (players) in $S$ act together. Therefore, from a Harsanyi-dividend perspective, we have the following analogous forms of Postulates~\ref{postulate_precision}, \ref{postulate_federation}, \ref{postulate_zero_discrepancy}, and Property~\ref{property_adaptation}: 

\textbf{Postulate 1h}. \textbf{\textit{Precision-h}}. The Harsanyi dividend of the $i$-th feature, $H(\{i\})$, shall be entirely attributed to the $i$-th feature, while it shall not be attributed to any other feature.

\textbf{Postulate 2h}. \textbf{\textit{Federation-h}}. Every Harsanyi dividend $H(S)$ shall be attributed only to the features in $S$.

\textbf{Postulate 3h}. \textbf{\textit{Zero-discrepancy-h}}. This postulate is identical to
Postulate~\ref{postulate_zero_discrepancy}.

\textbf{Property 1h}. \textbf{\textit{Adaptation-h}}. The proportion of $H(S)$ allocated to the $i$-th feature shall be tunable for every $S\subseteq N$ such that $|S|>1$ and $i\in S$.

Postulates 1h and 2h, together with Property 1h, follow directly
from Equation (\ref{eqn_taylorpoda_har_apdx}). In particular, for Postulate 3h, we have
\begin{equation}
\label{eqn_taylorpoda_federationh_apdx}
    \begin{aligned}
        f(\mathbf{b}) + \sum_{i\in N} \underset{(\text{TaylorPODA})}{a_i}
        &= f(\mathbf{b}) + \sum_{i\in N}(H(\{i\}) + \sum_{\substack{S\subseteq N\\|S|>1\\S\supseteq\{i\}}}\xi_{i, S}H(S))\\
        &= f(\mathbf{b}) + \sum_{i\in N}H(\{i\}) + \sum_{i\in N}\sum_{\substack{S\subseteq N\\|S|>1\\S\supseteq\{i\}}}\xi_{i, S}H(S)\\
        &= f(\mathbf{b}) + \sum_{i\in N}H(\{i\}) + \sum_{\substack{S\subseteq N\\|S|>1}}H(S)\\
        &= f(\mathbf{b}) + \sum_{\substack{S\subseteq N\\S\neq \varnothing}}H(S)\\
        &= f(\mathbf{x}).\\
    \end{aligned}
\end{equation}
Therefore, $f(\mathbf{b}) + \sum_{i\in N} a_i - f(\mathbf{x}) = 0$ holds, so that Postulate 3h is satisfied.

This completes the proof. \hfill $\blacksquare$

\section{Implementation details of the experiments}
\label{apdx_implementation}


\paragraph{Datasets.}

Six tabular datasets together with an image dataset are used for the experiments, all of which are publicly available. The $28\times28$ $\mathrm{MNIST}_{3,8}$ dataset is obtained by extracting all images of digits 3 and 8 from
the original \textit{MNIST} dataset. All datasets are shuffled, with 80\% of the samples in each dataset randomly selected for training. Performance evaluations are conducted using 100 hold-out samples randomly drawn from the remaining 20\%. For the classification tasks, explanations are generated with respect to the predicted score for the positive class. 

\paragraph{Setup for fair comparison of TaylorPODA and the other LA methods.}
The proposed TaylorPODA, together with the existing OCC-1, SHAP, and WeightedSHAP, can all be expressed within a common LA formulation and therefore share a similar computational procedure. To ensure a fair comparison, we configure the corresponding parameters and experimental procedures to exploit this shared computational structure. Specifically, since these LA methods rely on the masked outputs of the task models, they are configured to utilize a shared masked-output calculator
implemented via the \texttt{weightedshap.generate\_coalition\_function} subfunction provided by the WeightedSHAP package (\url{https://github.com/ykwon0407/WeightedSHAP}, used with the author's permission). For the qualitative experimental results presented in Figure~\ref{figure_mnist}, we used the \texttt{PermutationExplainer} from SHAP (version 0.44.0, MIT license) to generate Shapley-value-based image explanations. For LIME (version 0.2.0.1, BSD-2-Clause license), we utilized the \texttt{LimeImageExplainer} tailored for image data. 

\paragraph{Adopted utility functions.}
For completeness, the mathematical formulations of the utility functions adopted in our empirical analysis are presented. These utility functions quantify attribution quality under different evaluation settings, including Inclusion/Exclusion AUC for classification, Inclusion/Exclusion MSE for regression, and Inclusion/Exclusion AUP for both classification and regression. Higher values are preferred for Inclusion AUC, Exclusion MSE, and Exclusion AUP, whereas lower values are preferred for the remaining utility functions. For the MSE-based utilities, we follow the corresponding squared-error accumulation logic used in the empirical implementation; the averaging factor is omitted, since it does not affect the relative comparison or the optimization direction. Let $\mathcal{I}(m;\mathbf{a})$ denote the set of the top-$m$ features ranked by the absolute attribution values in $\mathbf{a}$, let $y(\mathbf{x})$ denote the corresponding ground-truth label, and let $\hat{y}_{\mathcal{I}(m;\mathbf{a})}(\mathbf{x})$ denote the predicted label obtained from the masked output with feature subset ${\mathcal{I}(m;\mathbf{a})}$. The corresponding utility functions are defined as follows.

Inclusion AUC:
\begin{equation}
\label{eqn_incauc}
\begin{aligned}
\underset{(\text{Inclusion AUC})}{u(\boldsymbol{\xi};f,\mathbf{x})} = \sum_{m=1}^{n}\mathbbm{1}\left[\hat{y}_{\mathcal{I}(m;\mathbf{a})}(\mathbf{x}) = y(\mathbf{x}) \right].
\end{aligned}
\end{equation}

Exclusion AUC:
\begin{equation}
\label{eqn_excauc}
\begin{aligned}
\underset{(\text{Exclusion AUC})}{u(\boldsymbol{\xi};f,\mathbf{x})} = \sum_{m=1}^{n}\mathbbm{1}\left[\hat{y}_{N\setminus \mathcal{I}(m;\mathbf{a})}(\mathbf{x}) = y(\mathbf{x}) \right].
\end{aligned}
\end{equation}

Inclusion MSE:
\begin{equation}
\label{eqn_incmse}
\begin{aligned}
\underset{(\text{Inclusion MSE})}{u(\boldsymbol{\xi};f,\mathbf{x})}
=\sum_{m=1}^{n}\left(f(\mathbf{x})-f_{\mathcal{I}(m;\mathbf{a})}(\mathbf{x})\right)^2.
\end{aligned}
\end{equation}

Exclusion MSE:
\begin{equation}
\label{eqn_excmse}
\begin{aligned}
\underset{(\text{Exclusion MSE})}{u(\boldsymbol{\xi};f,\mathbf{x})}
=\sum_{m=1}^{n}
\left(
f(\mathbf{x})-
f_{N\setminus \mathcal{I}(m;\mathbf{a})}(\mathbf{x})
\right)^2.
\end{aligned}
\end{equation}

Inclusion AUP:
\begin{equation}
\label{eqn_incaup}
\begin{aligned}
\underset{(\text{Inclusion AUP})}{u(\boldsymbol{\xi};f,\mathbf{x})}
=\sum_{m=1}^{n}\left|f(\mathbf{x})-f_{\mathcal{I}(m;\mathbf{a})}(\mathbf{x})\right|.
\end{aligned}
\end{equation}
Exclusion AUP:
\begin{equation}
\label{eqn_excaup}
\begin{aligned}
\underset{(\text{Exclusion AUP})}{u(\boldsymbol{\xi};f,\mathbf{x})}
=\sum_{m=1}^{n}
\left|
f(\mathbf{x})-
f_{N\setminus \mathcal{I}(m;\mathbf{a})}(\mathbf{x})
\right|.
\end{aligned}
\end{equation}

\section{Approximation of TaylorPODA}
\label{apdx_approximation}

As an initial step toward improving scalability and demonstrating the potential of TaylorPODA on high-dimensional datasets, we introduce a heuristic approximation (TaylorPODA-$c$) that preserves only terms associated with low-cardinality feature subsets:
\begin{equation}
\label{eqn_taylorpoda_c_apdx}
\begin{aligned}
    \underset{(\text{TaylorPODA})}{a_i}
    \approx\underset{(\text{TaylorPODA-c})}{a_i^{(c)}}
    =f(\mathbf{x})-f_{N\backslash\{i\}}(\mathbf{x})-\sum_{\substack{S\subseteq N\\1<|S|\leq c\\S\supseteq\{i\}}}(1-\xi_{i, S})H(S),
\end{aligned}
\end{equation}
where $c \in N$ with $c > 1$ denotes a cap on the cardinality of $S$. Given the cardinality cap $c$, the approximation error between the attribution results produced by TaylorPODA-c and the full TaylorPODA is given by
\begin{equation}
\label{eqn_delta}
\begin{aligned}
\Delta_i (c) &= \Bigg|\underset{(\text{TaylorPODA-c})}{a_i^{(c)}} - \underset{(\text{TaylorPODA})}{a_i}\Bigg|
= \Bigg|\sum_{\substack{S\subseteq N\\|S|>1\\S\supseteq\{i\}}}(1-\xi_{i, S})H(S) - \sum_{\substack{S\subseteq N\\1<|S|\leq c\\S\supseteq\{i\}}}(1-\xi_{i, S})H(S)\Bigg|\\
&= \Bigg|\sum_{\substack{S\subseteq N\\|S|>c\\S\supseteq\{i\}}}(1-\xi_{i, S})H(S)\Bigg|
\leq \sum_{\substack{S\subseteq N\\|S|>c\\S\supseteq\{i\}}}(1-\xi_{i, S})\Bigg|H(S)\Bigg|
\leq \sum_{\substack{S\subseteq N\\|S|>c\\S\supseteq\{i\}}}\Bigg|H(S)\Bigg|.
\end{aligned}
\end{equation}

For the illustrative image experiments, we set $c=2$ as an initial choice to improve scalability. Although this choice represents a relatively coarse approximation, the resulting performance remains competitive, as demonstrated in Section~\ref{ss_additional_evaluation}. By capping the cardinality of the feature subsets, this approximation approach aligns with the insights provided by existing studies~\citep{pmlr_v202_li23at,ren2024where} on the symbolic representations encoded by trained DNNs. Specifically, the Harsanyi dividend $H(S)$ has been empirically observed to exhibit sparse value patterns (under conditions characterizing a well-trained DNN), as described by the statement “the DNN will only encode a relatively small number of sparse interactions between input variables” \citep{ren2024where}, which implies that most $H(S)$ values are close to zero. In particular, previous studies~\citep{pmlr_v202_li23at,ren2024where} have consistently shown that $ |H(S)| $ remains below an empirical threshold of $ 0.05 \cdot \max_{S'}|H(S')| $ for \textbf{most} $ S \subseteq N$ with $|S|>1$, especially when $|S|\gg 1$. That is, even relatively low-order interactions suffice to capture the majority of attribution mass, and higher-order interactions contribute only marginally. 

\end{document}

%% file: math_commands.tex
\usepackage{amsmath,amsfonts,bm}

\def\eqref#1{equation~\ref{#1}}

\def\1{\bm{1}}

\DeclareMathAlphabet{\mathsfit}{\encodingdefault}{\sfdefault}{m}{sl}
\SetMathAlphabet{\mathsfit}{bold}{\encodingdefault}{\sfdefault}{bx}{n}



%% file: ref.bib
@String{Springer = "Springer-Verlag" }

@ArtifactSoftware{R,
    title = {R: A Language and Environment for Statistical Computing},
    author = {{R Core Team}},
    organization = {R Foundation for Statistical Computing},
    address = {Vienna, Austria},
    year = {2019},
    url = {https://www.R-project.org/},
}

@article{gregorutti2017correlation,
  title={Correlation and variable importance in random forests},
  author={Gregorutti, Baptiste and Michel, Bertrand and Saint-Pierre, Philippe},
  journal={Statistics and Computing},
  volume={27},
  number={3},
  pages={659--678},
  year={2017},
  publisher={Springer}
}

@article{fisher2019all,
  title={All models are wrong, but many are useful: Learning a variable's importance by studying an entire class of prediction models simultaneously},
  author={Fisher, Aaron and Rudin, Cynthia and Dominici, Francesca},
  journal={Journal of {Machine Learning} Research},
  volume={20},
  number={177},
  pages={1--81},
  year={2019}
}

@inproceedings{jethani2022fastshap,
    title={Fast{SHAP}: Real-Time {Shapley} Value Estimation},
    author={Neil Jethani and Mukund Sudarshan and Ian Connick Covert and Su-In Lee and Rajesh Ranganath},
    booktitle={International Conference on Learning Representations},
    year={2022},
    url={https://openreview.net/forum?id=Zq2G_VTV53T}
}

@article{lundberg2020local,
  title={From local explanations to global understanding with explainable {AI} for trees},
  author={Lundberg, Scott M and Erion, Gabriel and Chen, Hugh and DeGrave, Alex and Prutkin, Jordan M and Nair, Bala and Katz, Ronit and Himmelfarb, Jonathan and Bansal, Nisha and Lee, Su-In},
  journal={Nature Machine Intelligence},
  volume={2},
  number={1},
  pages={56--67},
  year={2020},
  publisher={Nature Publishing Group}
}

@inproceedings{pmlr_v70_sundararajan17a,
  title={Axiomatic attribution for deep networks},
  author={Sundararajan, Mukund and Taly, Ankur and Yan, Qiqi},
  booktitle={International Conference on Machine Learning},
  pages={3319--3328},
  year={2017},
  organization={PMLR}
}

@inproceedings{pmlr_v202_li23at,
  title={Does a neural network really encode symbolic concepts?},
  author={Li, Mingjie and Zhang, Quanshi},
  booktitle={International Conference on Machine Learning},
  pages={20452--20469},
  year={2023},
  organization={PMLR}
}

@inproceedings{ren2024where,
    title={Where We Have Arrived in Proving the Emergence of Sparse Interaction Primitives in {DNNs}},
    author={Qihan Ren and Jiayang Gao and Wen Shen and Quanshi Zhang},
    booktitle={The Twelfth International Conference on Learning Representations},
    year={2024}
}

@inproceedings{chen2016xgboost,
  title={{XGBoost}: A scalable tree boosting system},
  author={Chen, Tianqi and Guestrin, Carlos},
  booktitle={Proceedings of the 22nd {ACM} {SIGKDD} International Conference on Knowledge Discovery and Data Mining},
  pages={785--794},
  year={2016}
}

@inproceedings{sundararajan2020shapley,
  title={The {Shapley--Taylor Interaction Index}},
  author={Sundararajan, Mukund and Dhamdhere, Kedar and Agarwal, Ashish},
  booktitle={International Conference on Machine Learning},
  pages={9259--9268},
  year={2020},
  organization={PMLR}
}

@article{hart1989potential,
  title={Potential, value, and consistency},
  author={Hart, Sergiu and Mas-Colell, Andreu},
  journal={Econometrica: Journal of the Econometric Society},
  pages={589--614},
  year={1989},
  publisher={JSTOR}
}

@article{dubey1977probabilistic,
  title={Probabilistic values for games},
  author={Dubey, Pradeep and Weber, Robert J}, 
  journal={Cowles Foundation Discussion Papers},
  year={1977}
}

@article{frye2020asymmetric,
  title={Asymmetric {Shapley} values: incorporating causal knowledge into model-agnostic explainability},
  author={Frye, Christopher and Rowat, Colin and Feige, Ilya},
  journal={Advances in Neural Information Processing Systems},
  volume={33},
  pages={1229--1239},
  year={2020}
}

@inproceedings{watson2022rational,
  title={Rational {Shapley} values},
  author={Watson, David},
  booktitle={Proceedings of the 2022 {ACM} Conference on Fairness, Accountability, and Transparency},
  pages={1083--1094},
  year={2022}
}

@article{aas2021explaining,
  title={Explaining individual predictions when features are dependent: More accurate approximations to {Shapley} values},
  author={Aas, Kjersti and Jullum, Martin and L{\o}land, Anders},
  journal={Artificial Intelligence},
  volume={298},
  pages={103502},
  year={2021},
  publisher={Elsevier}
}

@article{goldstein2015peeking,
  title={Peeking inside the black box: Visualizing statistical learning with plots of individual conditional expectation},
  author={Goldstein, Alex and Kapelner, Adam and Bleich, Justin and Pitkin, Emil},
  journal={Journal of Computational and Graphical Statistics},
  volume={24},
  number={1},
  pages={44--65},
  year={2015},
  publisher={Taylor \& Francis}
}

@article{robnik2008explaining,
  title={Explaining classifications for individual instances},
  author={Robnik-{\v{S}}ikonja, Marko and Kononenko, Igor},
  journal={IEEE Transactions on Knowledge and Data Engineering},
  volume={20},
  number={5},
  pages={589--600},
  year={2008},
  publisher={IEEE}
}

@article{friedman2001greedy,
  title={Greedy function approximation: a gradient boosting machine},
  author={Friedman, Jerome H},
  journal={Annals of Statistics},
  pages={1189--1232},
  year={2001},
  publisher={JSTOR}
}

@book{ng2011dirichlet,
  title={Dirichlet and related distributions: Theory, methods and applications},
  author={Ng, Kai Wang and Tian, Guo-Liang and Tang, Man-Lai},
  year={2011},
  publisher={John Wiley \& Sons}
}

@inproceedings{lime,
  author    = {Marco Tulio Ribeiro and
               Sameer Singh and
               Carlos Guestrin},
  title     = {“{Why Should I Trust You?”: Explaining the Predictions of Any Classifier}},
  booktitle = {Proceedings of the 22nd {ACM} {SIGKDD} International Conference on
               {Knowledge Discovery and Data Mining}, San Francisco, CA, {USA}, August
               13-17, 2016},
  pages     = {1135--1144},
  year      = {2016},
}

@article{lecun1998mnist,
  title={Gradient-based learning applied to document recognition},
  author={LeCun, Yann and Bottou, L{\'e}on and Bengio, Yoshua and Haffner, Patrick},
  journal={Proceedings of the IEEE},
  volume={86},
  number={11},
  pages={2278--2324},
  year={1998},
  publisher={IEEE}
}

@article{harsanyi1982simplified,
  title={A simplified bargaining model for the n-person cooperative game},
  author={Harsanyi, John C},
  journal={Papers in game theory},
  pages={44--70},
  year={1982},
  publisher={Springer}
}

@inproceedings{zeiler2014visualizing,
  title={Visualizing and understanding convolutional networks},
  author={Zeiler, Matthew D and Fergus, Rob},
  booktitle={Computer Vision--ECCV 2014: 13th European Conference, Zurich, Switzerland, September 6-12, 2014, Proceedings, Part I 13},
  pages={818--833},
  year={2014},
  organization={Springer}
}

@article{deng2024unifying,
  title={Unifying fourteen post-hoc attribution methods with {Taylor} interactions},
  author={Deng, Huiqi and Zou, Na and Du, Mengnan and Chen, Weifu and Feng, Guocan and Yang, Ziwei and Li, Zheyang and Zhang, Quanshi},
  journal={IEEE Transactions on Pattern Analysis and Machine Intelligence},
  year={2024},
  publisher={IEEE}
}

@article{kwonWeightedSHAPAnalyzingImproving2022,
  title = {{WeightedSHAP}: Analyzing and improving {Shapley}-based feature attributions},
  shorttitle = {WeightedSHAP},
  author = {Kwon, Yongchan and Zou, James Y.},
  year = {2022},
  month = dec,
  journal = {Advances in Neural Information Processing Systems},
  volume = {35},
  pages = {34363--34376},
  urldate = {2024-04-19},
  langid = {english}
}

@article{lundberg2017unified,
  title={A unified approach to interpreting model predictions},
  author={Lundberg, Scott M and Lee, Su-In},
  journal={Advances in Neural Information Processing Systems},
  volume={30},
  year={2017}
}

@article{vstrumbelj2014explaining,
  title={Explaining prediction models and individual predictions with feature contributions},
  author={{\v{S}}trumbelj, Erik and Kononenko, Igor},
  journal={Knowledge and Information Systems},
  volume={41},
  pages={647--665},
  year={2014},
  publisher={Springer}
}

@article{shapley1953value,
  title={A value for n-person games},
  author={Shapley, Lloyd S},
  journal={Contribution to the Theory of Games},
  volume={2},
  year={1953}
}

@article{grabisch1999axiomatic,
  title={An axiomatic approach to the concept of interaction among players in cooperative games},
  author={Grabisch, Michel and Roubens, Marc},
  journal={International Journal of Game Theory},
  volume={28},
  number={4},
  pages={547--565},
  year={1999},
  publisher={Springer}
}

@article{jin2022explainable,
  title={Explainable deep learning in healthcare: A methodological survey from an attribution view},
  author={Jin, Di and Sergeeva, Elena and Weng, Wei-Hung and Chauhan, Geeticka and Szolovits, Peter},
  journal={WIREs Mechanisms of Disease},
  volume={14},
  number={3},
  pages={e1548},
  year={2022},
  publisher={Wiley Online Library}
}

@article{vcernevivciene2024explainable,
  title={Explainable {Artificial Intelligence} ({XAI}) in finance: a systematic literature review},
  author={{\v{C}}ernevi{\v{c}}ien{\.e}, Jurgita and Kaba{\v{s}}inskas, Audrius},
  journal={Artificial Intelligence Review},
  volume={57},
  number={8},
  pages={216},
  year={2024},
  publisher={Springer}
}

@article{machlev2022explainable,
  title={Explainable {Artificial Intelligence} ({XAI}) techniques for energy and power systems: Review, challenges and opportunities},
  author={Machlev, Ram and Heistrene, Leena and Perl, Michael and Levy, Kfir Yehuda and Belikov, Juri and Mannor, Shie and Levron, Yoash},
  journal={Energy and {AI}},
  volume={9},
  pages={100169},
  year={2022},
  publisher={Elsevier}
}

@article{salih2025perspective,
  title={A perspective on explainable artificial intelligence methods: {SHAP} and {LIME}},
  author={Salih, Ahmed M and Raisi-Estabragh, Zahra and Galazzo, Ilaria Boscolo and Radeva, Petia and Petersen, Steffen E and Lekadir, Karim and Menegaz, Gloria},
  journal={Advanced Intelligent Systems},
  volume={7},
  number={1},
  pages={2400304},
  year={2025},
  publisher={Wiley Online Library}
}

@misc{cancer,
  author       = {Wolberg, William},
  title        = {{Breast Cancer Wisconsin (Original)}},
  year         = {1990},
  howpublished = {UCI {Machine Learning} Repository},
  note         = {{DOI}: https://doi.org/10.24432/C5HP4Z}
}

@article{rice,
  title={Classification of Rice Varieties Using Artificial Intelligence Methods},
  author={Ilkay Cınar and Murat Koklu},
  journal={International Journal of Intelligent Systems and Applications in Engineering},
  year={2019},
  url={https://api.semanticscholar.org/CorpusID:208105752}
}

@misc{titanic,
    author = {Will Cukierski},
    title = {Titanic - {Machine Learning} from {Disaster}},
    year = {2012},
    howpublished = {\url{https://kaggle.com/competitions/titanic}},
    note = {Kaggle}
}

@misc{abalone,
  author       = {Nash, Warwick and Sellers, Tracy and Talbot, Simon and Cawthorn, Andrew and Ford, Wes},
  title        = {{Abalone}},
  year         = {1994},
  howpublished = {UCI {Machine Learning} Repository},
  note         = {{DOI}: https://doi.org/10.24432/C55C7W}
}

@article{california,
  author  = {Pace, R. Kelley and Barry, Ronald},
  title   = {Sparse Spatial Autoregressions},
  journal = {Statistics and Probability Letters},
  volume  = {33},
  number  = {3},
  pages   = {291--297},
  year    = {1997}
}

@misc{concrete,
  author       = {Yeh, I-Cheng},
  title        = {{Concrete Compressive Strength}},
  year         = {1998},
  howpublished = {UCI {Machine Learning} Repository},
  note         = {{DOI}: https://doi.org/10.24432/C5PK67}
}

@article{von2021transparency,
  title={Transparency and the black box problem: Why we do not trust {AI}},
  author={Von Eschenbach, Warren J},
  journal={Philosophy \& Technology},
  volume={34},
  number={4},
  pages={1607--1622},
  year={2021},
  publisher={Springer}
}

@inproceedings{dhanorkar2021needs,
  title={Who needs to know what, when?: Broadening the Explainable {AI} ({XAI}) Design Space by Looking at Explanations Across the {AI} Lifecycle},
  author={Dhanorkar, Shipi and Wolf, Christine T and Qian, Kun and Xu, Anbang and Popa, Lucian and Li, Yunyao},
  booktitle={Proceedings of the 2021 {ACM} Designing Interactive Systems Conference},
  pages={1591--1602},
  year={2021}
}
